\documentclass[10pt,twocolumn,letterpaper]{article}

\usepackage[table,dvipsnames]{xcolor}
\usepackage{iccv}
\usepackage{times}
\usepackage{graphicx}
\usepackage{amsmath}
\usepackage{amssymb}
\usepackage{enumitem}
\usepackage[export]{adjustbox}
\usepackage{array}
\usepackage{makecell}
\usepackage{color}

\usepackage{microtype}

\usepackage[,breaklinks=true,letterpaper=true,colorlinks,bookmarks=false]{hyperref}

 \iccvfinalcopy % *** Uncomment this line for the final submission

\def\iccvPaperID{353} % *** Enter the ICCV Paper ID here

\begin{document}

%%%%%%%%% TITLE
%\title{Recognition Beyond Photography: The \raisebox{-2.5em}{\includegraphics[width=3cm]{paper-figures/bam-logo.jpg}} Dataset\vspace{-2.5em}}
\title{BAM! The \emph{Behance Artistic Media} Dataset for Recognition Beyond Photography}
\author{
Michael J. Wilber$^{1,2}$
$\quad$
Chen Fang$^1$
$\quad$
Hailin Jin$^1$
$\quad$
Aaron Hertzmann$^1$
\\
John Collomosse$^1$
$\quad$
Serge Belongie$^2$
\\
$^1$ Adobe Research
$\quad$
$^2$ Cornell Tech
}
%\title{The Behance-Media Dataset: The fine art of fine tuning on fine art}

% Title ideas in [[file:~/notes/internship-2016.org::*Title%20ideas]]

\maketitle
%\thispagestyle{empty}

%%%%%%%%% ABSTRACT
\begin{abstract}
Computer vision systems are designed to work well within the context of everyday photography. However, artists often render the world around them in ways that do not resemble photographs. Artwork produced by people is not constrained to mimic the physical world, making it more challenging for machines to recognize.

This work is a step toward teaching machines how to categorize images in ways that are valuable to humans. First, we collect a large-scale dataset of contemporary artwork from Behance, a website containing millions of portfolios from professional and commercial artists. We annotate Behance imagery with rich attribute labels for content, emotions, and artistic media.
Furthermore, we carry out baseline experiments to show the value of this dataset for artistic style prediction, for improving the generality of existing object classifiers, and for the study of visual domain adaptation.
%Finally, we show how to use this dataset to train models for artistic style prediction tasks and to improve the generality of existing object classifiers.
We believe our Behance Artistic Media dataset will be a good starting point for researchers wishing to study artistic imagery and relevant problems. \textbf{This dataset can be found at \url{https://bam-dataset.org/}}

\end{abstract}

\begin{figure}[t]
  \centering
\newcommand{\rcol}[1]{\rotatebox[origin=l]{90}{\parbox[c]{50pt}{\centering \textbf{#1}}}}
\newcommand{\imgexample}[5]{%
\begin{picture}(50,0)%
\put(0,6){\includegraphics[width=0.20\linewidth]{image-cache/image-examples/#3}}%
\put(8,0){\scriptsize \textcolor{#1}{Score: #4}}%
%\put(10,0){\scriptsize \textcolor{#2}{Ours: #5}}%
\end{picture}}
\setlength{\tabcolsep}{0.0em}
\begin{tabular}{lcccc}
& \textbf{Bicycle} & \textbf{Bird} & \textbf{Cat} & \textbf{Dog} \vspace{0.3em}\\%
\rcol{Photography}&
\imgexample{OliveGreen}{OliveGreen}{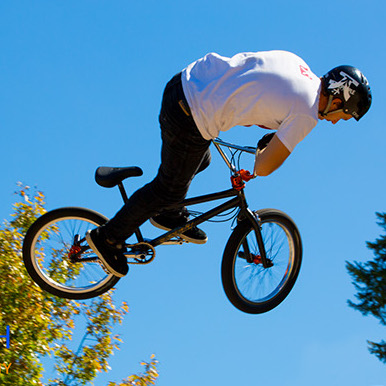}{0.99}{0.99}&%
\imgexample{OliveGreen}{OliveGreen}{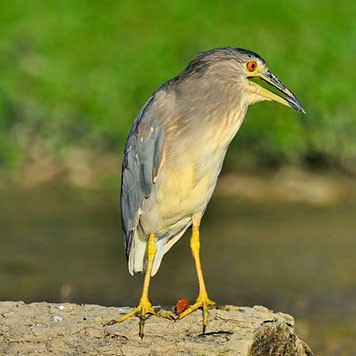}{0.99}{0.99}&%
\imgexample{OliveGreen}{OliveGreen}{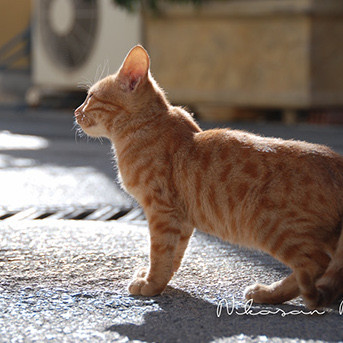}{0.99}{0.99}&%
\imgexample{OliveGreen}{OliveGreen}{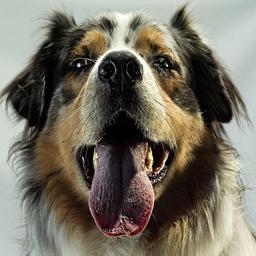}{0.99}{0.99}%
\\\rcol{Comic}&
\imgexample{Maroon}{OliveGreen}{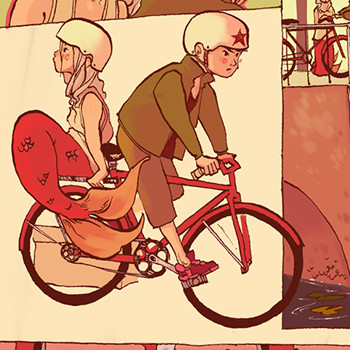}{0.01}{0.99}&%
%\imgexample{Maroon}{OliveGreen}{content-bicycle-media-comic/0_01539-11221803-content-bicycle-media-comic.jpg}{}{}&%
\imgexample{Maroon}{OliveGreen}{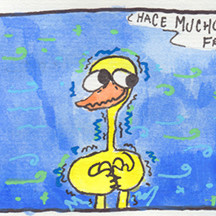}{0.01}{0.93}&%
\imgexample{Maroon}{OliveGreen}{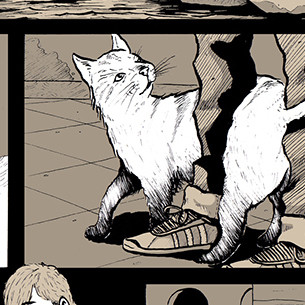}{0.01}{0.97}&%
%\imgexample{Maroon}{OliveGreen}{content-dog-media-comic/0_01063-171603183.jpg}{}{}%
%\imgexample{Maroon}{OliveGreen}{content-dog-media-comic/0_01194-52651011.jpg}{}{}%
%\imgexample{Maroon}{OliveGreen}{content-dog-media-comic/0_02940-53456277.jpg}{}{}%
%\imgexample{Maroon}{OliveGreen}{content-dog-media-comic/0_03360-67280107.jpg}{}{}%
\imgexample{Maroon}{OliveGreen}{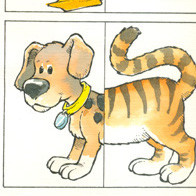}{0.34}{0.99}%
\\\rcol{Pencil}&
%\imgexample{Maroon}{OliveGreen}{content-bicycle-media-graphite/0_01335-142356899-content-bicycle-media-graphite.jpg}{}{}&%
\imgexample{Maroon}{OliveGreen}{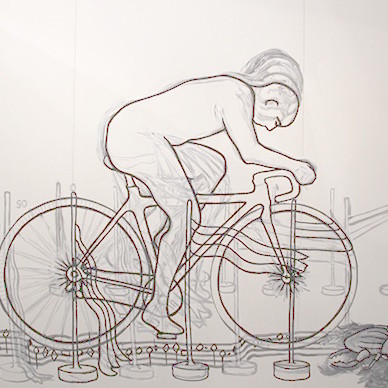}{0.02}{0.99}&%
%\imgexample{Maroon}{OliveGreen}{content-bicycle-media-graphite/0_02838-168162625-content-bicycle-media-graphite.jpg}{0.28}{0.99}&%
\imgexample{Maroon}{OliveGreen}{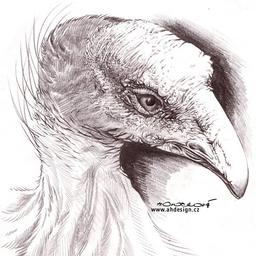}{0.01}{0.99}&%
%\imgexample{Maroon}{OliveGreen}{content-bird-media-graphite/0_01028-30151613-content-bird-media-graphite.jpg}{0.01}{0.99}&%
%\imgexample{Maroon}{OliveGreen}{content-cat-media-graphite/0_01055-115830803.jpg}{}{}&%
%\imgexample{Maroon}{OliveGreen}{content-cat-media-graphite/0_01219-153335035.jpg}{}{}&%
\imgexample{Maroon}{OliveGreen}{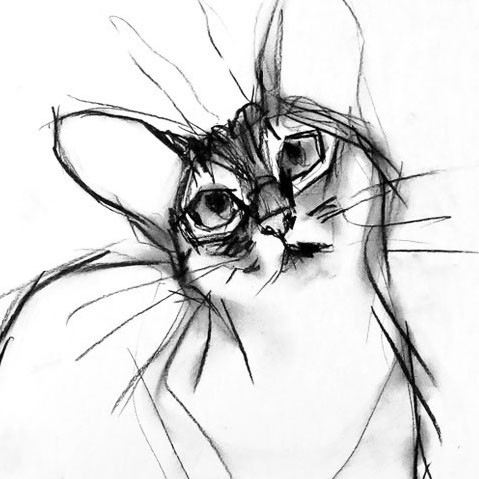}{0.03}{0.99}&%
\imgexample{Maroon}{OliveGreen}{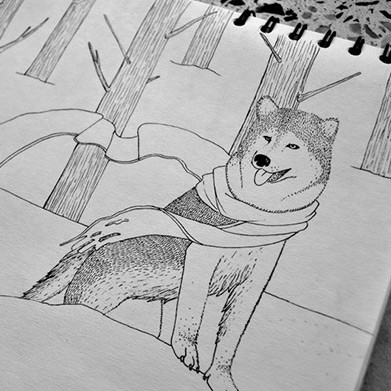}{0.01}{0.96}%
\\\rcol{Oil paint}&
%\imgexample{Maroon}{OliveGreen}{content-bicycle-media-oilpaint/0_01485-133578509-content-bicycle-media-oilpaint.jpg}{}{}&%
%\imgexample{Maroon}{OliveGreen}{content-bicycle-media-oilpaint/0_01711-4072993-content-bicycle-media-oilpaint.jpg}{}{}&%
%\imgexample{Maroon}{OliveGreen}{content-bicycle-media-oilpaint/0_06635-142950685-content-bicycle-media-oilpaint.jpg}{}{}&%
\imgexample{Maroon}{OliveGreen}{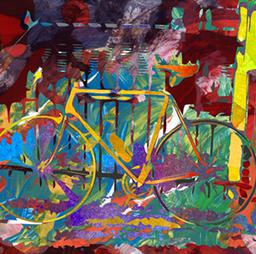}{0.07}{0.88}&%
%\imgexample{Maroon}{OliveGreen}{content-bicycle-media-oilpaint/0_12225-80555525-content-bicycle-media-oilpaint.jpg}{}{}&%
\imgexample{Maroon}{OliveGreen}{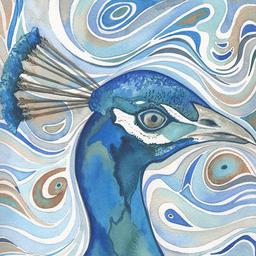}{0.00}{0.97}&%
\imgexample{Maroon}{OliveGreen}{content-cat-media-oilpaint/0_01073-152269745.jpg}{0.01}{0.99}&%
%\imgexample{Maroon}{OliveGreen}{content-cat-media-oilpaint/0_01485-144372041.jpg}{}{}&%
%\imgexample{Maroon}{OliveGreen}{content-dog-media-oilpaint/0_01350-124716835.jpg}{}{}&%
\imgexample{Maroon}{OliveGreen}{content-dog-media-oilpaint/0_01706-120788479.jpg}{0.02}{0.99}%
%\imgexample{Maroon}{OliveGreen}{content-dog-media-oilpaint/0_01738-53450187.jpg}{}{}&%
\\\rcol{Vector art}&
\imgexample{Maroon}{OliveGreen}{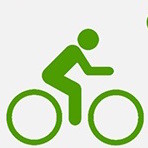}{0.00}{0.99}&%
\imgexample{Maroon}{OliveGreen}{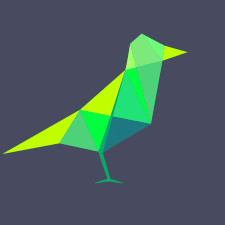}{0.01}{0.99}&%
%\imgexample{Maroon}{OliveGreen}{content-bird-media-vectorart/0_01169-105069701-content-bird-media-vectorart.jpg}{}{}&%
\imgexample{Maroon}{OliveGreen}{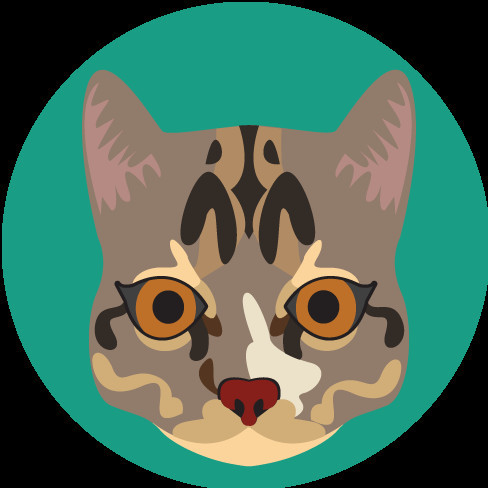}{0.01}{0.99}&%
%\imgexample{Maroon}{OliveGreen}{content-dog-media-vectorart/0_01063-171603183.jpg}{0.01}{0.97}%
\imgexample{Maroon}{OliveGreen}{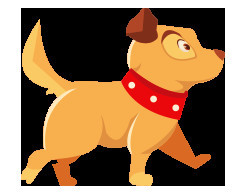}{0.01}{0.97}%
%\imgexample{Maroon}{OliveGreen}{content-dog-media-vectorart/0_07994-162929035.jpg}{}{}&%
\\\rcol{Watercolor}&
%\imgexample{Maroon}{OliveGreen}{content-bicycle-media-watercolor/0_16405-126896521-content-bicycle-media-watercolor.jpg}{}{}&%
\imgexample{Maroon}{OliveGreen}{content-bicycle-media-watercolor/0_28063-173598125.jpeg}{0.28}{0.99}&%
%\imgexample{Maroon}{OliveGreen}{content-bird-media-watercolor/0_01008-107355415.jpg}{}{}&%
%\imgexample{Maroon}{OliveGreen}{content-bird-media-watercolor/0_01228-49740177.jpg}{}{}&%
\imgexample{Maroon}{OliveGreen}{content-bird-media-watercolor/0_01542-105067007.jpg}{0.02}{0.99}&%
\imgexample{Maroon}{OliveGreen}{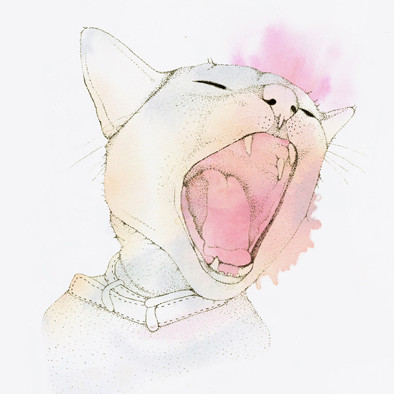}{0.01}{0.99}&%
%\imgexample{Maroon}{OliveGreen}{content-cat-media-watercolor/0_01485-144372041.jpg}{}{}&%
%\imgexample{Maroon}{OliveGreen}{content-dog-media-watercolor/0_01051-134421197.jpg}{}{}%
%\imgexample{Maroon}{OliveGreen}{content-dog-media-watercolor/0_01350-124716835.jpg}{}{}%
\imgexample{Maroon}{OliveGreen}{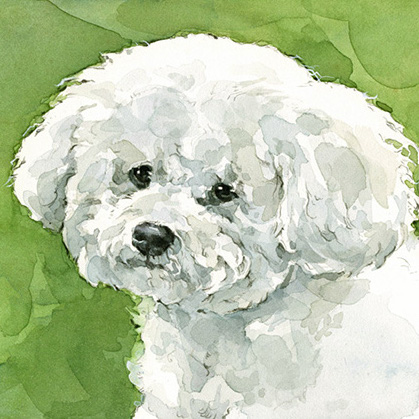}{0.02}{0.99}%
\end{tabular}
  \caption{
    \label{fig:content-style}
State of the art object detectors such as SSD trained on Pascal VOC can reliably detect objects in everyday photographs (top row), but do not generalize to other kinds of artistic media (see scores under each image). In this work, \textbf{we create a large-scale artistic dataset} spanning a breadth of styles, media, and emotions. We can use this dataset to improve the generality of object classifiers---our object classifier's scores are above 0.95 for all these images.
}
\end{figure}
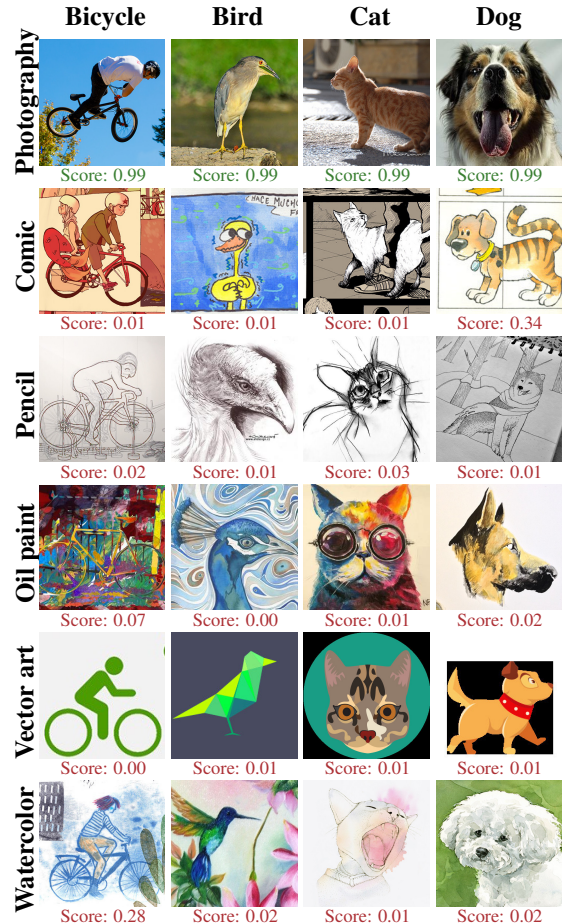

\section{Introduction}

%\setlength{\textfloatsep}{10.0pt minus 4.0pt}
%\setlength{\floatsep}{10.0pt minus 4.0pt}
%\setlength{\textfloatsep}{10.0pt plus 2.0pt minus 4.0pt}

 % Machines are slaves to their datasets ... ... Must free ourselves from the shackles of natural representations! ... ...

\begin{quote}
\emph{``Art is an effort to create, beside the real world, a more humane world.'' -- Andr\'e Maurois}
% \emph{``If you don't want a generation of robots, fund the arts!'' -- Cath Crowley, Graffiti Moon}
% Art is the best possible window into another culture.
% ...
% "Life imitates Art far more than Art imitates Life" -- Oscar Wilde
\end{quote}
% Here's a fun possible intro:
% CV is advancing on three fronts:
% - More categories, working well on increasingly fine-grained data;
% - More accuracy,
% -
% We want to extend this to the "frontier of representation," pushing beyond just photographs.
% Serge's suggested paragraph
Recent advances in Computer Vision have yielded accuracy rivaling that of humans on a variety of object recognition tasks. However, most work in this space is focused on understanding \emph{photographic imagery} of everyday scenes. For example, the widely-used COCO dataset \cite{mscoco} was created by ``gathering images of complex everyday scenes containing common objects in their natural context.'' Outside of everyday photography, there exists a diverse, relatively unexplored space of \emph{artistic imagery}, offering depictions of the world \emph{as reinterpreted through artwork}. Besides being culturally valuable, artwork spans broad styles that are not found in everyday photography and thus are not available to current machine vision systems. For example, current object classifiers trained on ImageNet and Pascal VOC are frequently unable to recognize objects when they are depicted in artistic media (Fig.~\ref{fig:content-style}).
Modeling artistic imagery can increase the generality of computer vision models by pushing beyond the limitations of photographic datasets.
%Further, besides being inherently and culturally valuable to humans, artistic understanding is important for several applications, including HCI, document analysis, image retrieval and affective computing, and machine-assisted art and design.

In this work, we create a large-scale artistic style dataset from Behance, a website containing millions of portfolios from professional and commercial artists. Content on Behance spans several industries and fields, ranging from creative direction to fine art to technical diagrams to graffiti to concept design. Behance does not aim to be a historical archive of classic art; rather, we start from Behance because it represents a broad cross-section of contemporary art and design.

Our overall goal is to create a dataset that researchers can use as a testbed for studying artistic representations across different artistic media. %to teach machines to understand and categorize artistic images in ways that are valuable to humans.
This is important because existing artistic datasets are too small or are focused on classical artwork, ignoring the different styles found in contemporary digital artwork.
%This is not an easy task to define because artwork can be categorized in many different ways. No matter the scope of the project, exhaustively annotating a dataset of artwork also requires human expertise, which can be costly at large scale.
To solidify the scope of the problem, we choose to explore three different facets of high-level image categorization: object categories, artistic media, and emotions.
These artistic facets are attractive for several reasons: they are readily understood by non-experts, they can describe a broad range of contemporary artwork, and they are not apparent from current photographic datasets.

% Teaching machines about different styles is important because
%
We keep the following goals in mind when deciding which attributes to annotate. For object categories, we wish to annotate objects that may be drawn in many different visual styles, collecting fewer visually distinct categories but increasing the density (instances per category) and breadth of representation.
ImageNet and COCO, for example, contain rich fine-grained object annotations, but these datasets are focused on everyday photos and cover a narrow range of artistic representation.
For media attributes, we wish to annotate pictures rendered with all kinds of professional media: pencil sketches, computer-aided vector illustration, watercolor, and so on.
%Datasets such as Wikipaintings aim to be a historically accurate archive of classical artwork across genres and time periods, but do not aim to capture a wide breadth of contemporary artistic media. \aaron{this sentence looks really out-of-place. Why is it here?}
%
Finally, emotion is an important categorization facet that is relatively unexplored by current approaches. %Designers wishing to locate imagery used

There are several challenges, including annotating millions of images in a scalable way, defining a categorization vocabulary that represents the style and content of Behance, and using this resource to study how well object recognition systems generalize to unseen domains.
According to our quality tests, the precision of the labels in our dataset is 90\%, which is reasonable for such a large dataset without consortium level funding.

\textbf{Our contributions} are twofold:
\begin{itemize}[topsep=0pt,itemsep=-1ex,partopsep=1ex,parsep=1ex]
\item \textbf{A large-scale dataset}, the Behance Artistic Media Dataset, containing almost 65 million images and quality assurrance thresholds. We also create an \textbf{expert-defined vocabulary} of binary artistic attributes that spans the broad spectrum of artistic styles and content represented in Behance. This dataset can be found at \url{https://bam-dataset.org/} upon release late Spring 2017.
%\item An \textbf{iterative label bootstrapping algorithm} that allows us to annotate this dataset at low cost while satisfying quality guarantees by focusing the crowd's attention on the most worthwhile images to label.
\item An investigation of the \textbf{representation gap} between objects in everyday ImageNet photographs and objects rendered in artistic media on Behance. We also explore how models trained on one medium can transfer that performance to unseen media in a \textbf{domain adaptation setting}.
To investigate aesthetics and art styles, we compare performance of different kinds of features in predicting emotion and media and show how Behance Artistic Media can be used to improve style classification tasks on other datasets.
Finally, we briefly investigate \textbf{style-aware image search}, showing how our dataset can be used to search for images based on their content, media, or emotion.
\end{itemize}
We believe this dataset will provide a starting foundation for researchers who wish to expand the horizon of machine vision to the rich domain of artwark.

% Note to reviewers: We will release the dataset upon publication. We plan to increase the number of attributes upon release.

\section{Related Work}
\begin{table}
\scalebox{0.6}{%
\begin{tabular}{llll}
& Size & Scope & Annotations \\
\hline
A-SUN \cite{Patterson2012SUNAD} & 0.014m & Photos of scenes & Objects, context \\
Behance-2M (Private) \cite{Fang2015CollaborativeFL} & 1.9m & Contemporary artwork & User/View behavior \\
Recognizing Image Style \cite{Karayev2014RecognizingIS} & 0.16m & Photos, paintings & Art genre, photo techniques\\
AVA \cite{Murray2012AVAAL} & 0.25m & Photos & Aesthetics, content, style\\
%The ontology paper ... \cite{Jou2015} & 7.4m & Photos only & Adj/Noun pairs\\
Visual sentiment ontology \cite{Borth2013} & 0.31m & Photos, videos & Adj/Noun pairs \\
OpenImages \cite{openimages} & 9.2m & Photos & Content labels \\
\textbf{Behance Artistic Media} & \textbf{65m} & \textbf{Contemporary artwork} & \textbf{Emotion, Media, Objects}
\end{tabular}
}
\caption{\label{tab:datasets}A comparison of several related datasets. Our Behance Artistic Media dataset is much larger than the others and includes a broad range of contemporary artwork.}
\end{table}

%In our work, we collect a bank of image attributes to describe and categorize artistic works.
Attributes and other mid-level representations~\cite{NunoSemanticMultinomials,Farhadi2009} have a long and rich history in vision. %Two seminal works that introduce attributes are the ``semantic multinomials'' of Rasiwasia and Vasconcelos~\cite{NunoSemanticMultinomials}, which lift images into a semantic space useful for performing visual searches, and the work by Farhadi \etal~\cite{Farhadi2009}, which use human-describable attributes to perform zero-shot learning of new objects.
%Attributes have been applied to face recognition via the work of Kumar \etal~\cite{Kumar2011DescribableVA} later extended by Scheirer \etal~\cite{Scheirer2012MultiattributeSC} and scene understanding by Patterson \etal~\cite{Patterson2012SUNAD}, Redi \etal~\cite{Redi2012EnhancingSF}, and others.
%
Attributes have been applied to aesthetics and other artistic qualities, usually with a focus on photography. For instance, Obrador \etal~\cite{Obrador2012TowardsCA}, Dhar \etal~\cite{Dhar2011HighLD}, and Murray \etal~\cite{Murray2012AVAAL} collect descriptive attributes such as interestingness, symmetry, light exposure, and depth of field. Work by Peng \etal~\cite{peng2016}, You \etal~\cite{You2016BuildingAL}, Jou \etal~\cite{Jou2015}, and Borth \etal~\cite{Borth2013} study emotional attributes in photographs. %Other work such as Jou \etal~\cite{Jou2015} and Borth \etal~\cite{Borth2013} use emotions to build ontologies of Adjective/Noun pairs to describe images.
% **** Emotion
% http://chenlab.ece.cornell.edu/Publication/Kuan-Chuan/ICIP16_EmotionROI.pdf
% \cite{peng2016}
% - Ekman's 6 basic emotions: Anger, disgust, joy, fear, sadness, surprise
% - Studies emotions ROI on photographs
Others describe image style not in attributes, but in terms of low-level feature correlations as in work done by Gatys \etal~\cite{Gatys2015}, Lin \etal~\cite{Lin_2016_CVPR}, and others. We are more concerned about high-level image categorization than low-level texture transfer. %The application studied in Gatys' work is transferring texture from one image to another, but we argue there is more to artistic style than low-level texture transfer. We are more concerned about high-level image categorization.

Ours is not the only dataset focused on artwork. We compare related artistic datasets in Tab.~\ref{tab:datasets}. Most are focused exclusively on everyday photographs~\cite{Murray2012AVAAL,Patterson2012SUNAD,Borth2013}, but some~\cite{Karayev2014RecognizingIS,Crowley14a,Ginosar2015} include classical paintings. Likewise, Ginosar \etal~\cite{Ginosar2015} discuss person detection in cubist art. The work of Fang \etal~\cite{Fang2015CollaborativeFL} also studies Behance imagery, but does not collect descriptive attributes. Recently, Google released the ``Open Images`` dataset~\cite{openimages} containing some media-related labels including ``comics'', ``watercolor paint'', ``graffiti'', etc. However, it is unclear how the quality of the labeling was evaluated. Each of these labels contain less than 400 human-verified images and there are no labels that capture emotions. %Open Images may contain some artistic imagery, but that is not its focus.
% **** Related Datasets
% Show off the above table.
% - \cite{Murray2012AVAAL} AVA dataset
% - \cite{Patterson2012SUNAD} SUN Attribute dataset
% - \cite{Karayev2014RecognizingIS} Recognizing image style set
% - The large-scale visual sentiment ontology
%- OpenImages Watercolor paint, Painting. We should be able to tell the difference between Behance and Open Images. It would be great to know a bit about Open Images’s tag distribution, i.e., whether there are many artistic tgas, and how many images are associated with these tags. And of course, how it is collected is important too.
%
Our work is most similar in spirit to Karayev \etal~\cite{Karayev2014RecognizingIS}, which studies photographic image style. They collect annotations for photographic techniques, composition, genre, and mood on Flickr images, as well as a set of classical painting genres on Wikipaintings. Our focus is on non-photorealistic contemporary art. %, which is also covered by Fang \etal~\cite{Fang2015CollaborativeFL}. Fang \etal's work trains a style prediction network to predict image ``pseudoclasses,'' which are clusters of images that encompass consistent styles according to user behavior. Our approach is explicit: we directly annotate semantically meaningful attributes from that feature space.
To our knowledge, our work is the first work seeking to release a large-scale dataset of a broad range of contemporary artwork with emotion, media, and content annotations.

\section{The Behance Media Dataset}
\begin{figure}[t]
\centering{\textbf{\footnotesize Random images from projects with tag ``Cat'':}} \\%
\includegraphics[width=0.13\linewidth,cfbox=red 0.5pt 0pt]{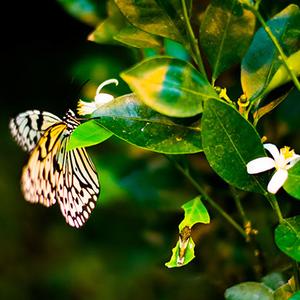}%
\includegraphics[width=0.13\linewidth,cfbox=green 0.5pt 0pt]{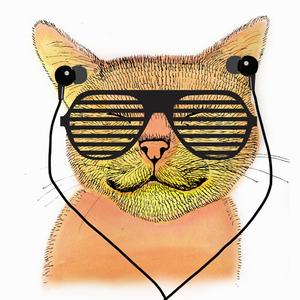}%
\includegraphics[width=0.13\linewidth,cfbox=red 0.5pt 0pt]{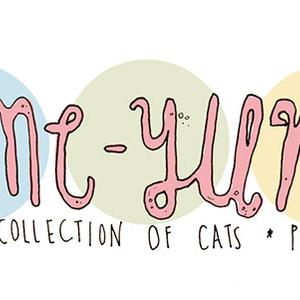}%
\includegraphics[width=0.13\linewidth,cfbox=green 0.5pt 0pt]{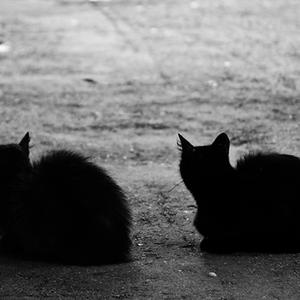}%
\includegraphics[width=0.13\linewidth,cfbox=red 0.5pt 0pt]{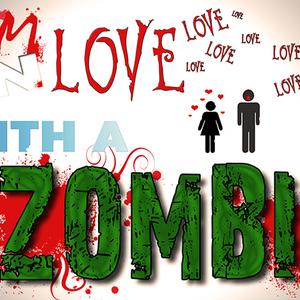}%
\includegraphics[width=0.13\linewidth,cfbox=red 0.5pt 0pt]{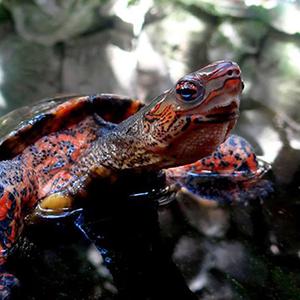}%
\includegraphics[width=0.13\linewidth,cfbox=red 0.5pt 0pt]{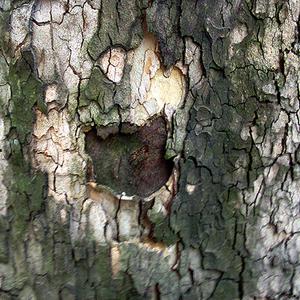}%
\\%
\centering{\textbf{\footnotesize Top classifications for ``Cat'' tag classifier:}} \\%
\includegraphics[width=0.13\linewidth,cfbox=red 0.5pt 0pt]{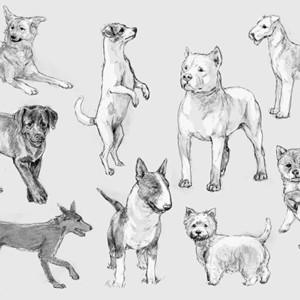}%
\includegraphics[width=0.13\linewidth,cfbox=red 0.5pt 0pt]{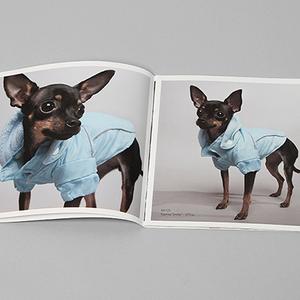}%
\includegraphics[width=0.13\linewidth,cfbox=red 0.5pt 0pt]{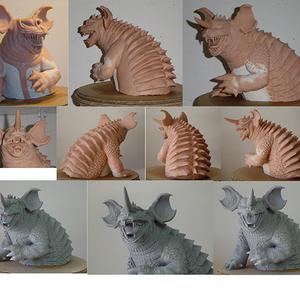}%
\includegraphics[width=0.13\linewidth,cfbox=red 0.5pt 0pt]{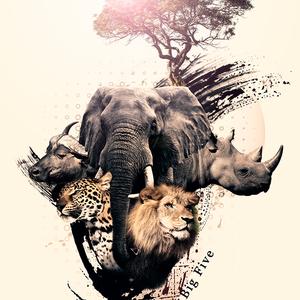}%
\includegraphics[width=0.13\linewidth,cfbox=red 0.5pt 0pt]{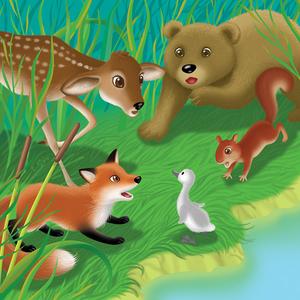}%
\includegraphics[width=0.13\linewidth,cfbox=green 0.5pt 0pt]{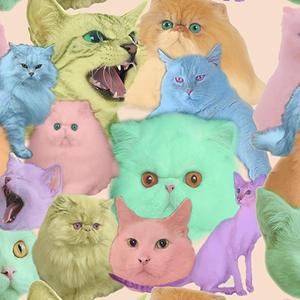}%
\includegraphics[width=0.13\linewidth,cfbox=green 0.5pt 0pt]{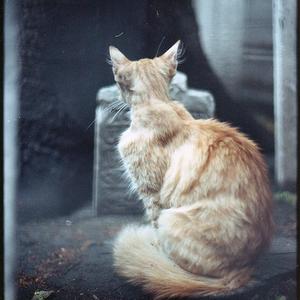}%
% \\%
% \includegraphics[width=0.15\linewidth]{image-cache/51012385.jpg}%
% \includegraphics[width=0.15\linewidth]{image-cache/132796375.jpg}%
% \includegraphics[width=0.15\linewidth]{image-cache/65001045.jpg}%
% \includegraphics[width=0.15\linewidth]{image-cache/166185935.jpg}%
% \includegraphics[width=0.15\linewidth]{image-cache/95177561.jpg}%
  \caption{\label{fig:tag-only-comparison}Top: Sampling of images within projects with the ``Cat'' tag. Projects with the ``Cat'' tag may contain other animals (1), title cards (3,5), or unrelated pictures (5,6). Bottom: Top classifications from a classifier trained to distinguish the ``Cat'' tag. Images are more related, but this tends to learn many small animals. The precision of cats in the top 100-scoring images is only 36\%.}
\end{figure}

Our dataset is built from \url{http://behance.net}, a portfolio website for professional and commercial artists. Behance contains over ten million projects and 65 million  images. Images on Behance are grouped into Projects, the fundamental unit of categorization. Each Project is associated with metadata, including a title, description, and several noisy user-supplied tags.

\begin{figure}[t]
    \newcommand{\tabletitle}[1]{{\textsf{\scriptsize \raisebox{1.5em}{#1}}}}
\setlength{\tabcolsep}{0.2em}
\hspace{-2em}
\setlength{\extrarowheight}{-14pt}
\renewcommand{\arraystretch}{0.8}
\scalebox{0.985}{
  \begin{tabular}{rc}
    & {\scriptsize \textbf{Content}} \\
 \tabletitle{bicycle} &
\includegraphics[width=0.11\linewidth]{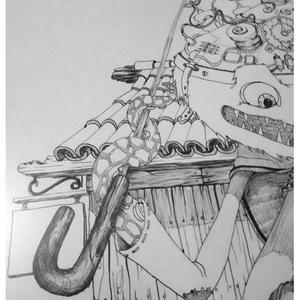}%
\includegraphics[width=0.11\linewidth]{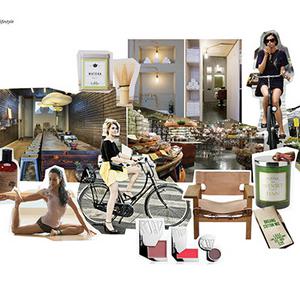}%
\includegraphics[width=0.11\linewidth]{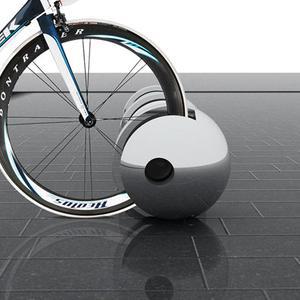}%
\includegraphics[width=0.11\linewidth]{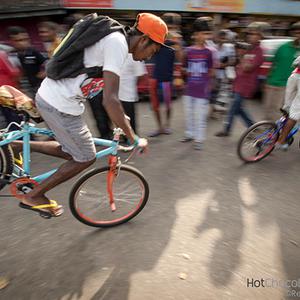}%
\includegraphics[width=0.11\linewidth]{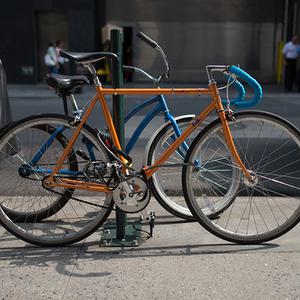}%
\includegraphics[width=0.11\linewidth]{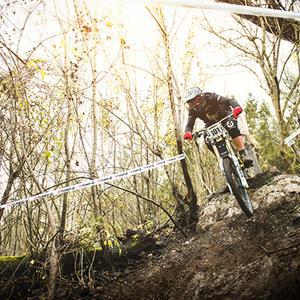}%
\includegraphics[width=0.11\linewidth]{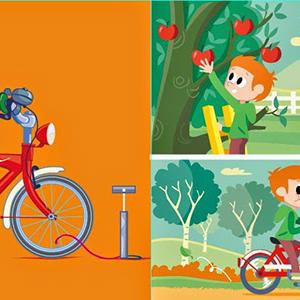}%
\includegraphics[width=0.11\linewidth]{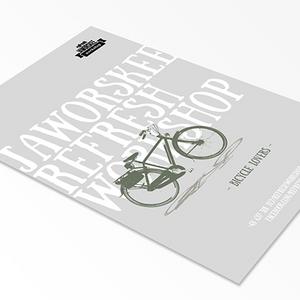}%
\\
\tabletitle{bird} &
\includegraphics[width=0.11\linewidth]{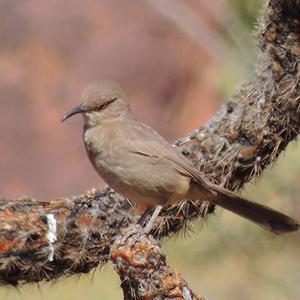}%
\includegraphics[width=0.11\linewidth]{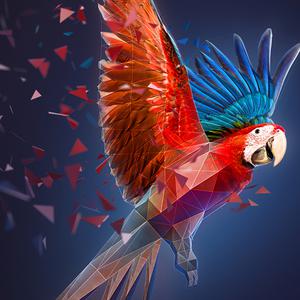}%
\includegraphics[width=0.11\linewidth]{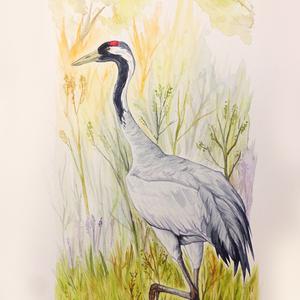}%
\includegraphics[width=0.11\linewidth]{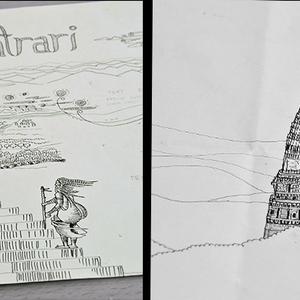}%
\includegraphics[width=0.11\linewidth]{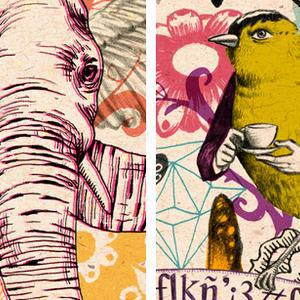}%
\includegraphics[width=0.11\linewidth]{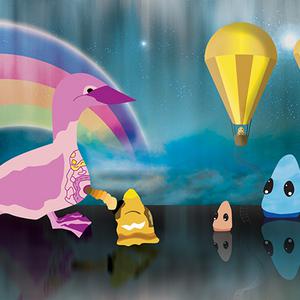}%
\includegraphics[width=0.11\linewidth]{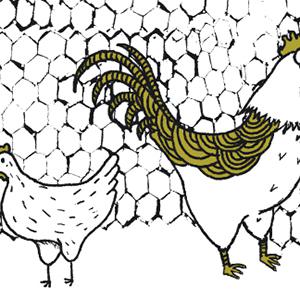}%
\includegraphics[width=0.11\linewidth]{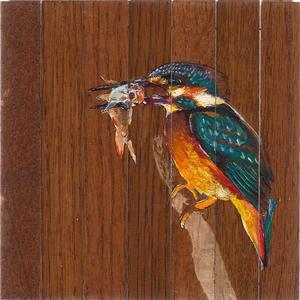}%
\\
\tabletitle{building} &
\includegraphics[width=0.11\linewidth]{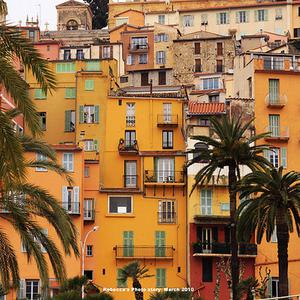}%
\includegraphics[width=0.11\linewidth]{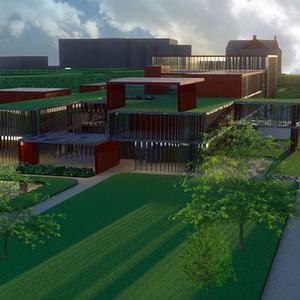}%
\includegraphics[width=0.11\linewidth]{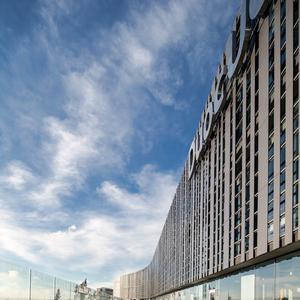}%
\includegraphics[width=0.11\linewidth]{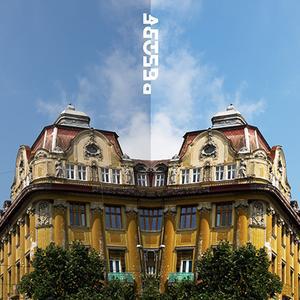}%
\includegraphics[width=0.11\linewidth]{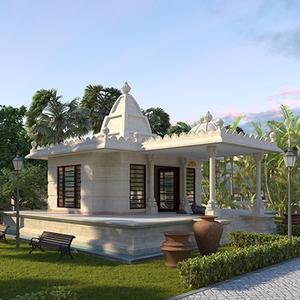}%
\includegraphics[width=0.11\linewidth]{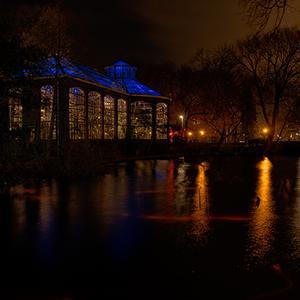}%
\includegraphics[width=0.11\linewidth]{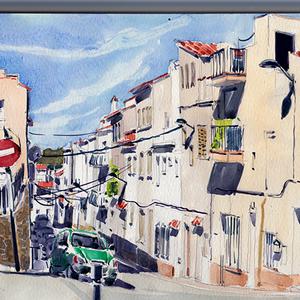}%
\includegraphics[width=0.11\linewidth]{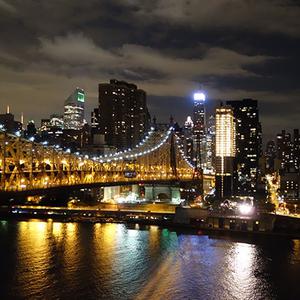}%
\\
\tabletitle{cars} &
\includegraphics[width=0.11\linewidth]{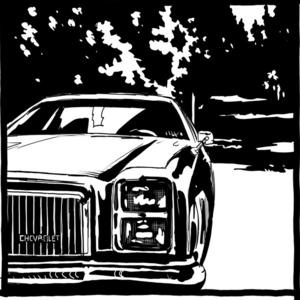}%
\includegraphics[width=0.11\linewidth]{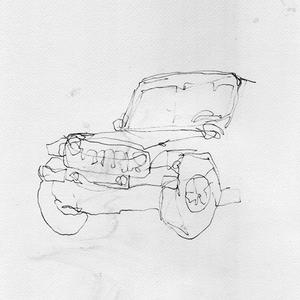}%
\includegraphics[width=0.11\linewidth]{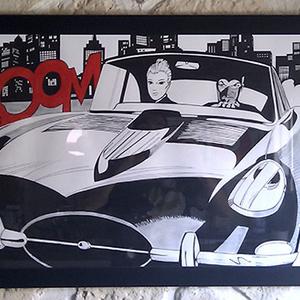}%
\includegraphics[width=0.11\linewidth]{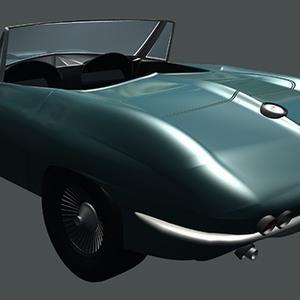}%
\includegraphics[width=0.11\linewidth]{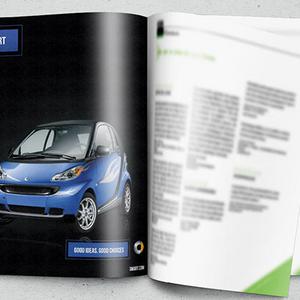}%
\includegraphics[width=0.11\linewidth]{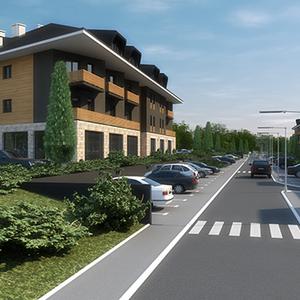}%
\includegraphics[width=0.11\linewidth]{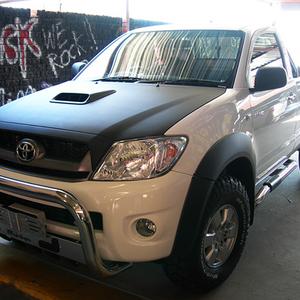}%
\includegraphics[width=0.11\linewidth]{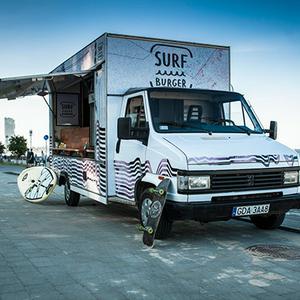}%
\\
\tabletitle{cat} &
\includegraphics[width=0.11\linewidth]{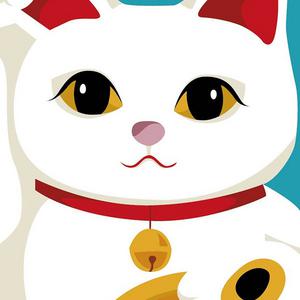}%
\includegraphics[width=0.11\linewidth]{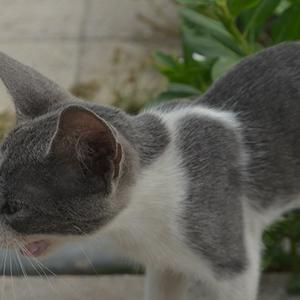}%
\includegraphics[width=0.11\linewidth]{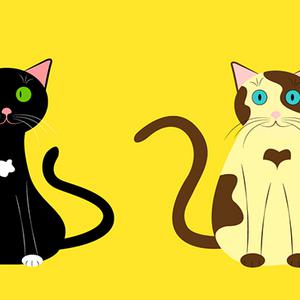}%
\includegraphics[width=0.11\linewidth]{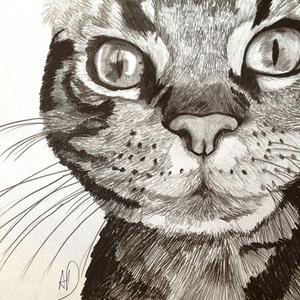}%
\includegraphics[width=0.11\linewidth]{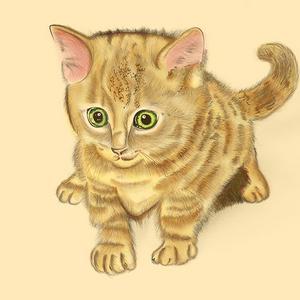}%
\includegraphics[width=0.11\linewidth]{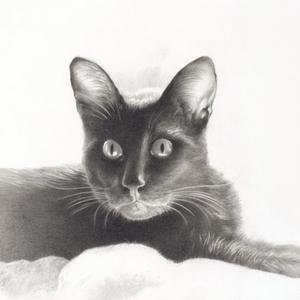}%
\includegraphics[width=0.11\linewidth]{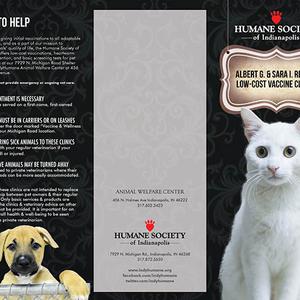}%
\includegraphics[width=0.11\linewidth]{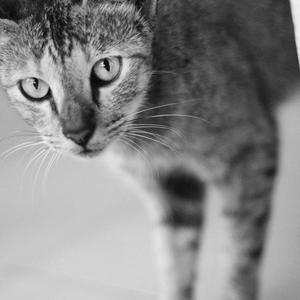}%
\\
\tabletitle{dog} &
\includegraphics[width=0.11\linewidth]{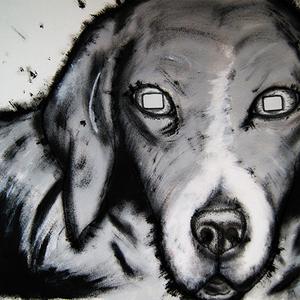}%
\includegraphics[width=0.11\linewidth]{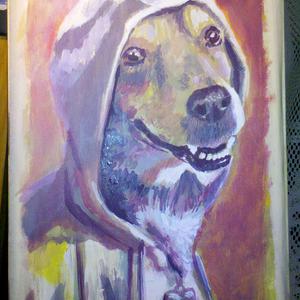}%
\includegraphics[width=0.11\linewidth]{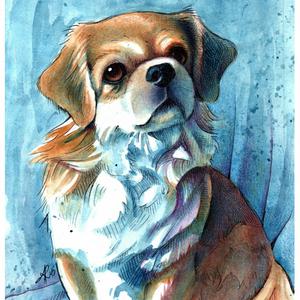}%
\includegraphics[width=0.11\linewidth]{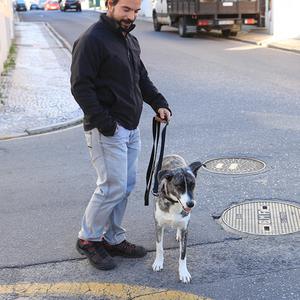}%
\includegraphics[width=0.11\linewidth]{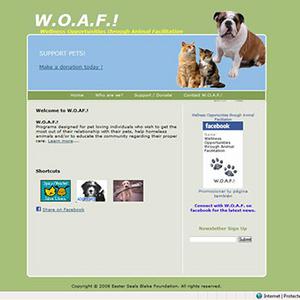}%
\includegraphics[width=0.11\linewidth]{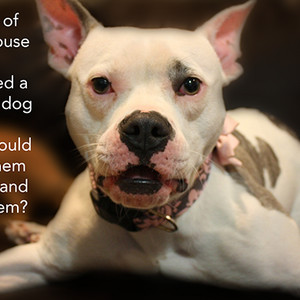}%
\includegraphics[width=0.11\linewidth]{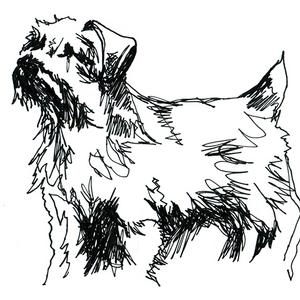}%
\includegraphics[width=0.11\linewidth]{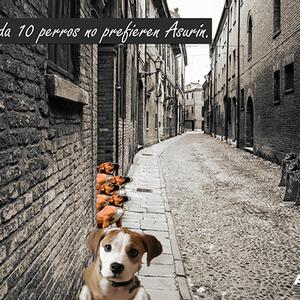}%
\\
\tabletitle{flower} &
\includegraphics[width=0.11\linewidth]{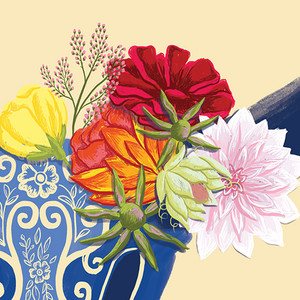}%
\includegraphics[width=0.11\linewidth]{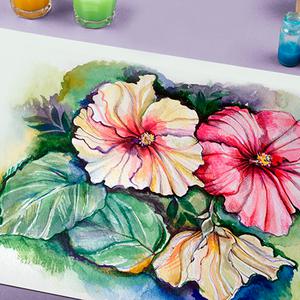}%
\includegraphics[width=0.11\linewidth]{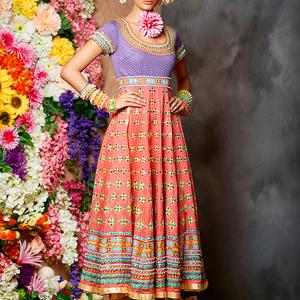}%
\includegraphics[width=0.11\linewidth]{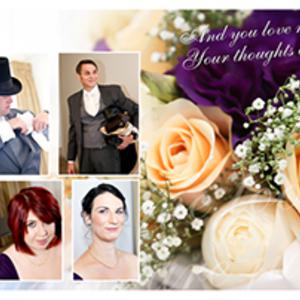}%
\includegraphics[width=0.11\linewidth]{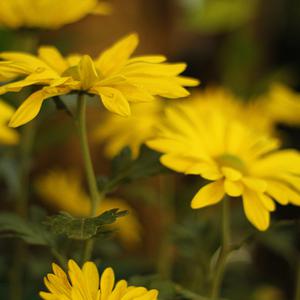}%
\includegraphics[width=0.11\linewidth]{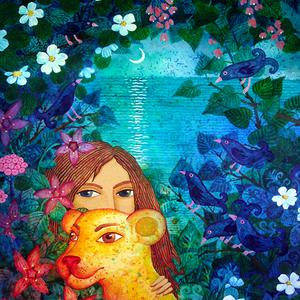}%
\includegraphics[width=0.11\linewidth]{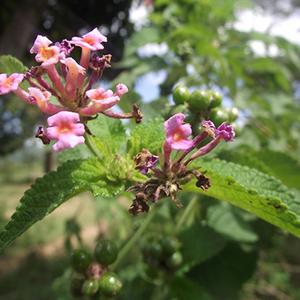}%
\includegraphics[width=0.11\linewidth]{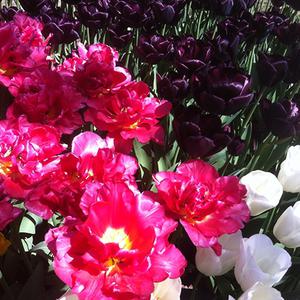}%
\\
\tabletitle{people} &
\includegraphics[width=0.11\linewidth]{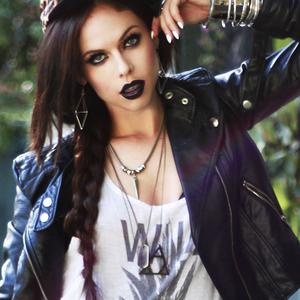}%
\includegraphics[width=0.11\linewidth]{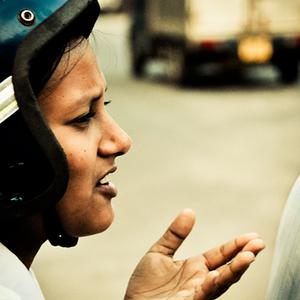}%
\includegraphics[width=0.11\linewidth]{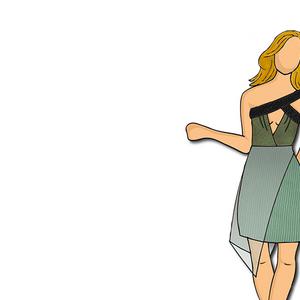}%
\includegraphics[width=0.11\linewidth]{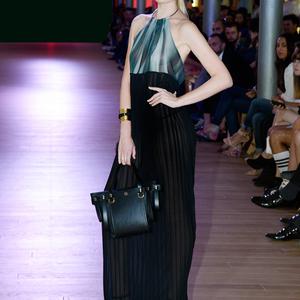}%
\includegraphics[width=0.11\linewidth]{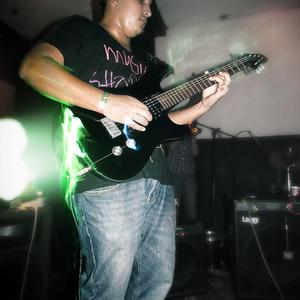}%
\includegraphics[width=0.11\linewidth]{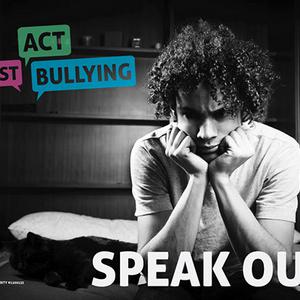}%
\includegraphics[width=0.11\linewidth]{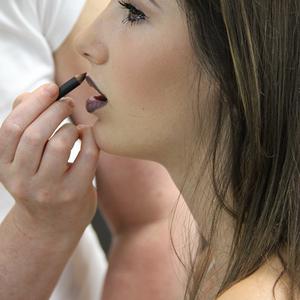}%
\includegraphics[width=0.11\linewidth]{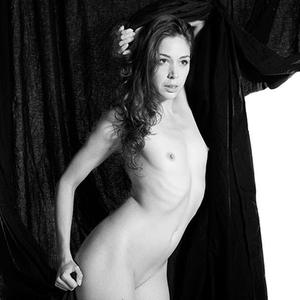}%
\\
\tabletitle{tree} &
\includegraphics[width=0.11\linewidth]{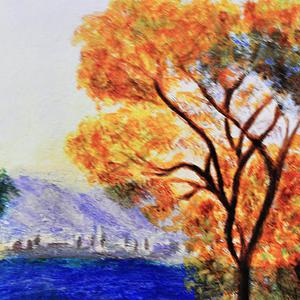}%
\includegraphics[width=0.11\linewidth]{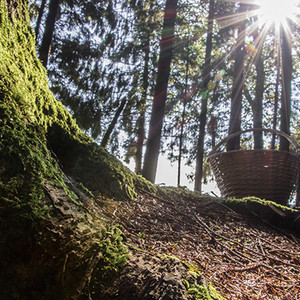}%
\includegraphics[width=0.11\linewidth]{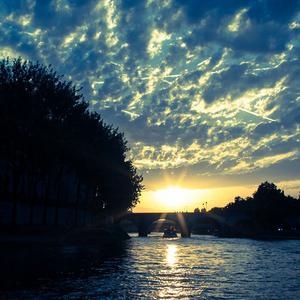}%
\includegraphics[width=0.11\linewidth]{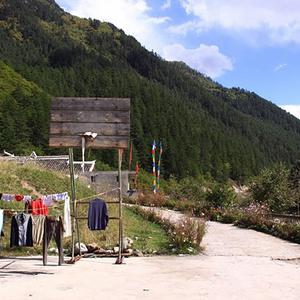}%
\includegraphics[width=0.11\linewidth]{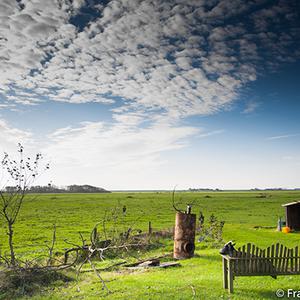}%
\includegraphics[width=0.11\linewidth]{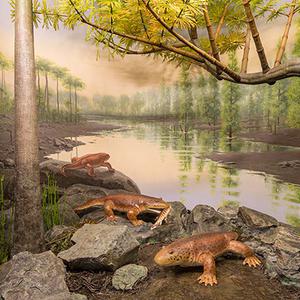}%
\includegraphics[width=0.11\linewidth]{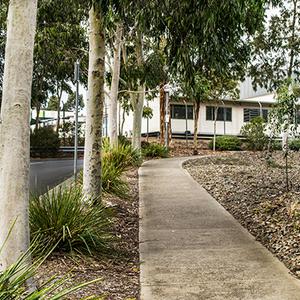}%
\includegraphics[width=0.11\linewidth]{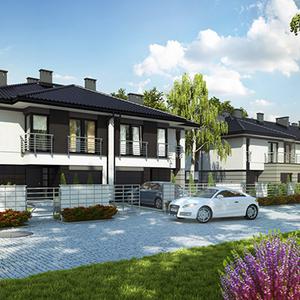}%
\\
    & {\scriptsize \textbf{Emotion}} \\
\tabletitle{gloomy} &
\includegraphics[width=0.11\linewidth]{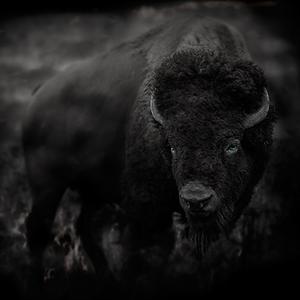}%
\includegraphics[width=0.11\linewidth]{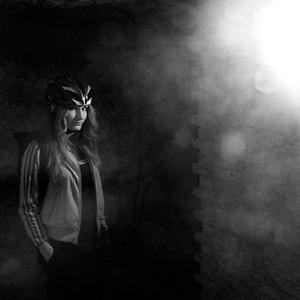}%
\includegraphics[width=0.11\linewidth]{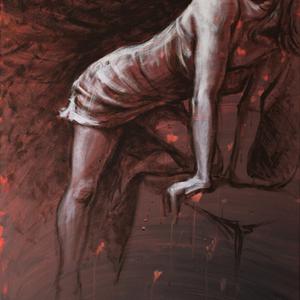}%
\includegraphics[width=0.11\linewidth]{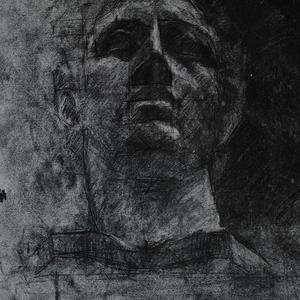}%
\includegraphics[width=0.11\linewidth]{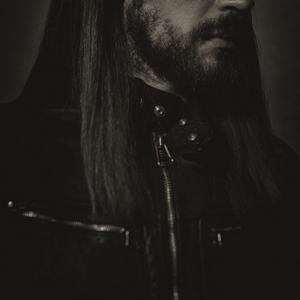}%
\includegraphics[width=0.11\linewidth]{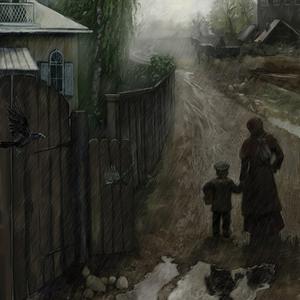}%
\includegraphics[width=0.11\linewidth]{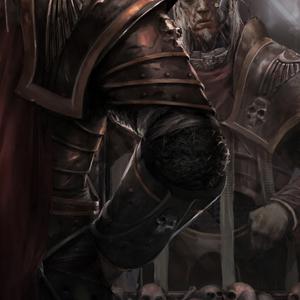}%
\includegraphics[width=0.11\linewidth]{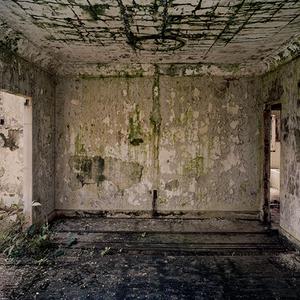}%
\\
\tabletitle{happy} &
\includegraphics[width=0.11\linewidth]{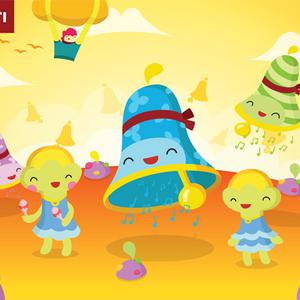}%
\includegraphics[width=0.11\linewidth]{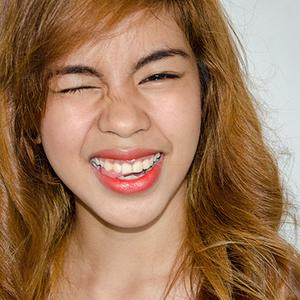}%
\includegraphics[width=0.11\linewidth]{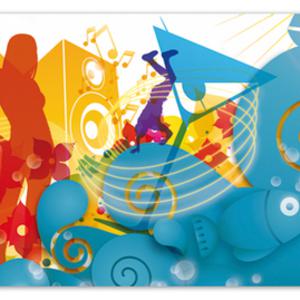}%
\includegraphics[width=0.11\linewidth]{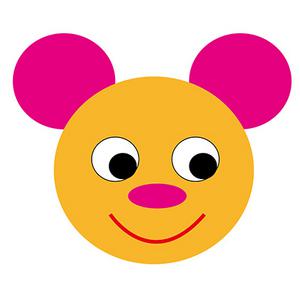}%
\includegraphics[width=0.11\linewidth]{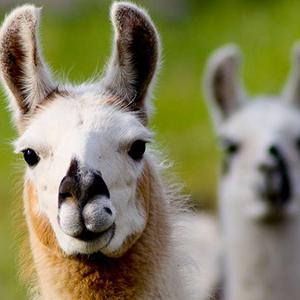}%
\includegraphics[width=0.11\linewidth]{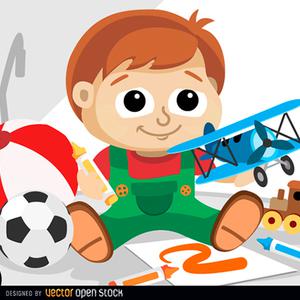}%
\includegraphics[width=0.11\linewidth]{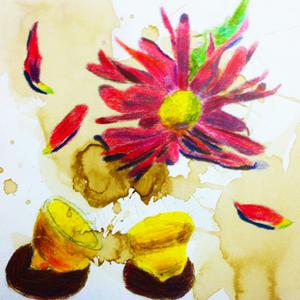}%
\includegraphics[width=0.11\linewidth]{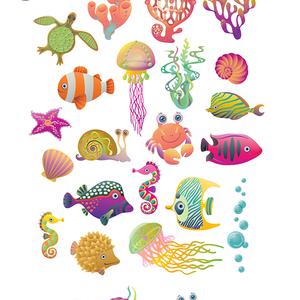}%
\\
\tabletitle{peaceful} &
\includegraphics[width=0.11\linewidth]{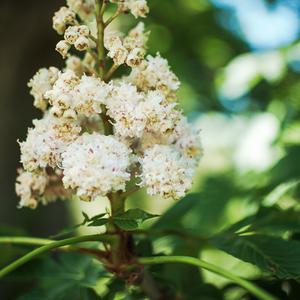}%
\includegraphics[width=0.11\linewidth]{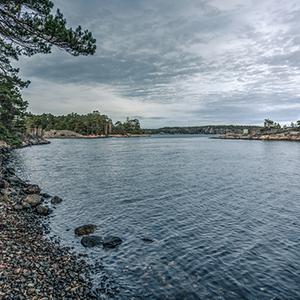}%
\includegraphics[width=0.11\linewidth]{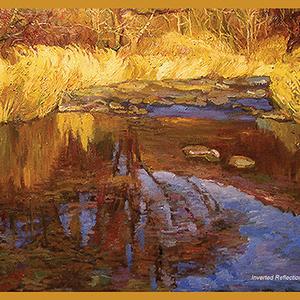}%
\includegraphics[width=0.11\linewidth]{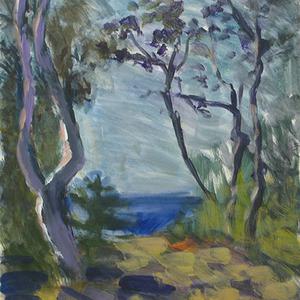}%
\includegraphics[width=0.11\linewidth]{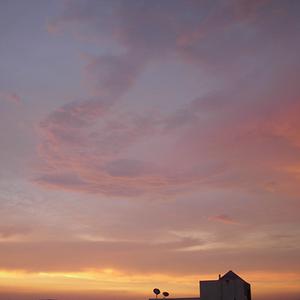}%
\includegraphics[width=0.11\linewidth]{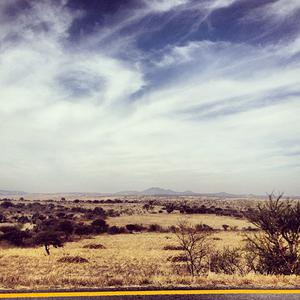}%
\includegraphics[width=0.11\linewidth]{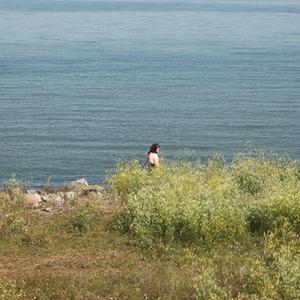}%
\includegraphics[width=0.11\linewidth]{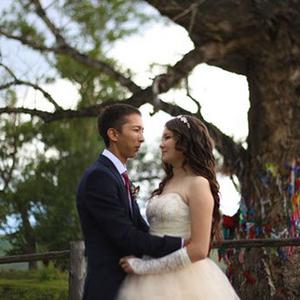}%
\\
\tabletitle{scary} &
\includegraphics[width=0.11\linewidth]{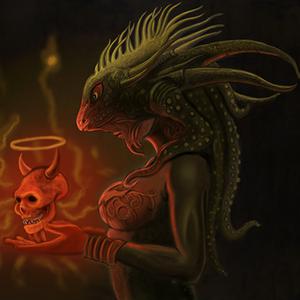}%
\includegraphics[width=0.11\linewidth]{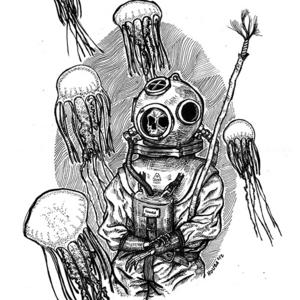}%
\includegraphics[width=0.11\linewidth]{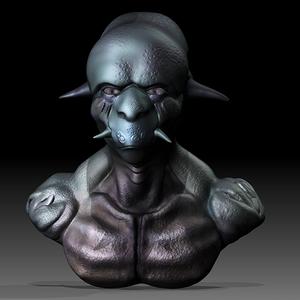}%
\includegraphics[width=0.11\linewidth]{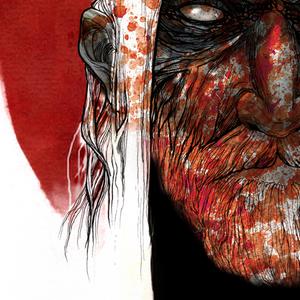}%
\includegraphics[width=0.11\linewidth]{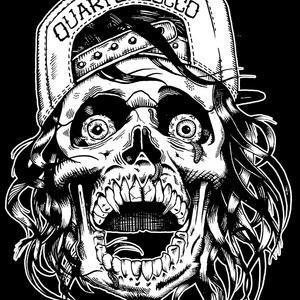}%
\includegraphics[width=0.11\linewidth]{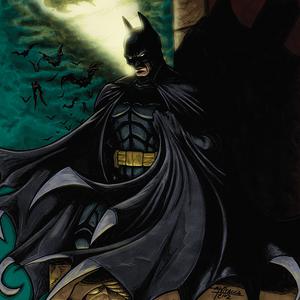}%
\includegraphics[width=0.11\linewidth]{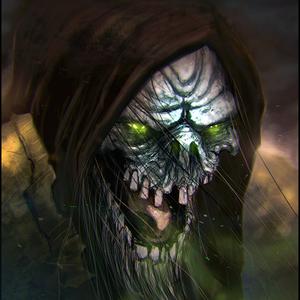}%
\includegraphics[width=0.11\linewidth]{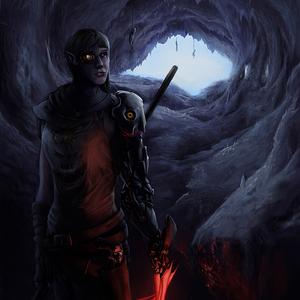}%
\\
    & {\scriptsize \textbf{Media}} \\
\tabletitle{3d} &
\includegraphics[width=0.11\linewidth]{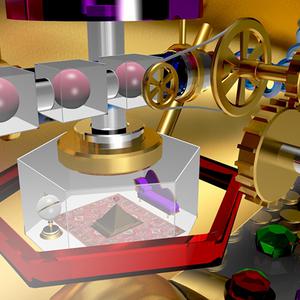}%
\includegraphics[width=0.11\linewidth]{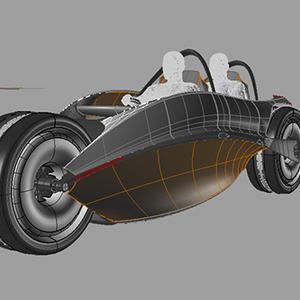}%
\includegraphics[width=0.11\linewidth]{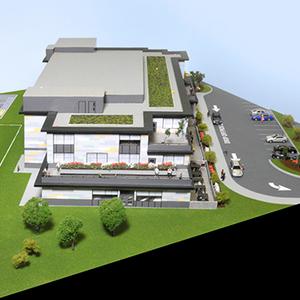}%
\includegraphics[width=0.11\linewidth]{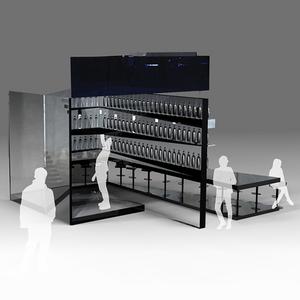}%
\includegraphics[width=0.11\linewidth]{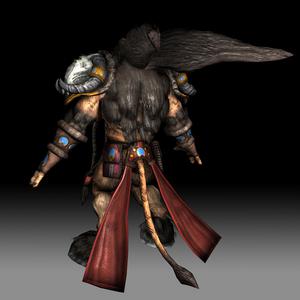}%
\includegraphics[width=0.11\linewidth]{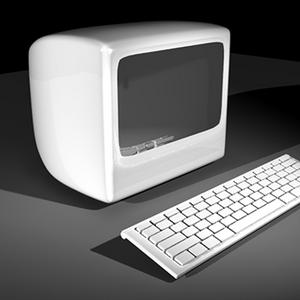}%
\includegraphics[width=0.11\linewidth]{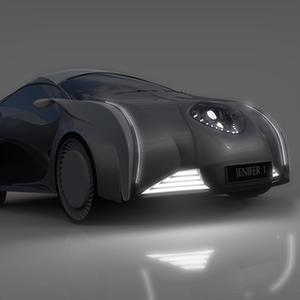}%
\includegraphics[width=0.11\linewidth]{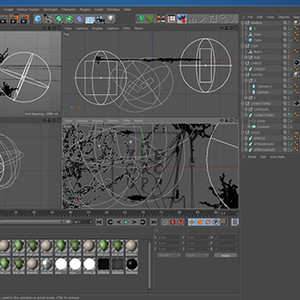}%
\\
\tabletitle{comic} &
\includegraphics[width=0.11\linewidth]{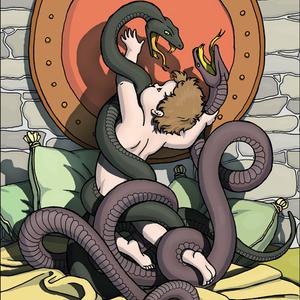}%
\includegraphics[width=0.11\linewidth]{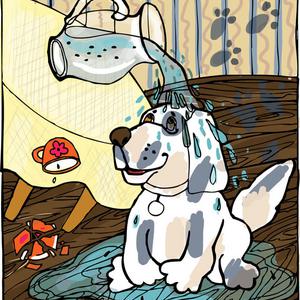}%
\includegraphics[width=0.11\linewidth]{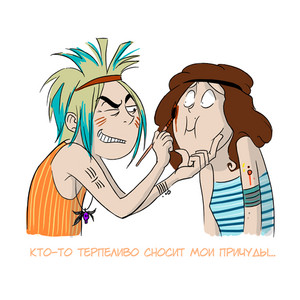}%
\includegraphics[width=0.11\linewidth]{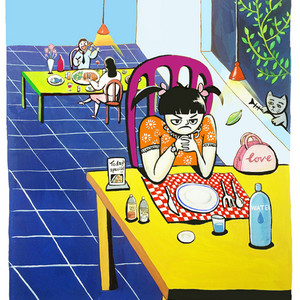}%
\includegraphics[width=0.11\linewidth]{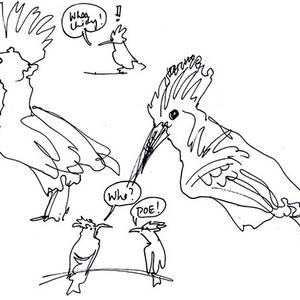}%
\includegraphics[width=0.11\linewidth]{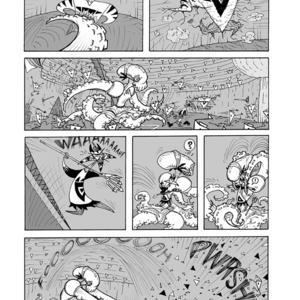}%
\includegraphics[width=0.11\linewidth]{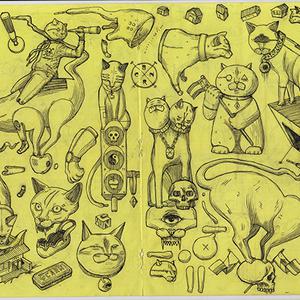}%
\includegraphics[width=0.11\linewidth]{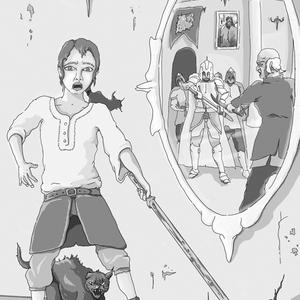}%
\\
\tabletitle{graphite} &
\includegraphics[width=0.11\linewidth]{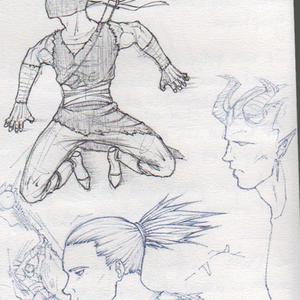}%
\includegraphics[width=0.11\linewidth]{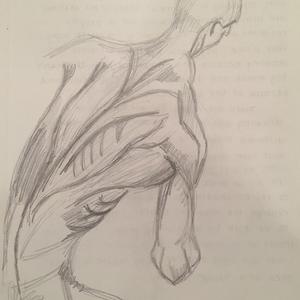}%
\includegraphics[width=0.11\linewidth]{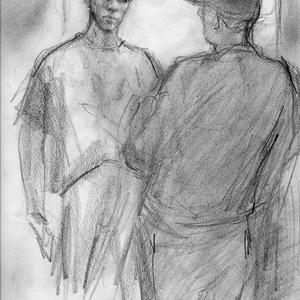}%
\includegraphics[width=0.11\linewidth]{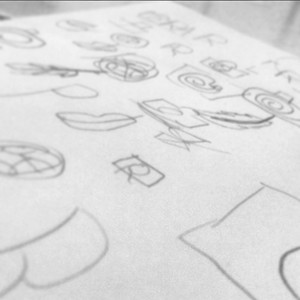}%
\includegraphics[width=0.11\linewidth]{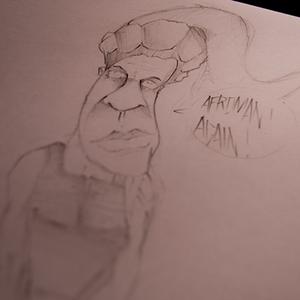}%
\includegraphics[width=0.11\linewidth]{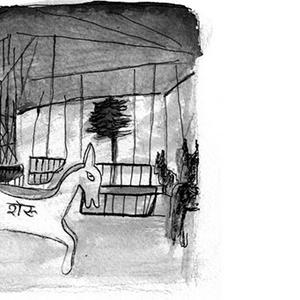}%
\includegraphics[width=0.11\linewidth]{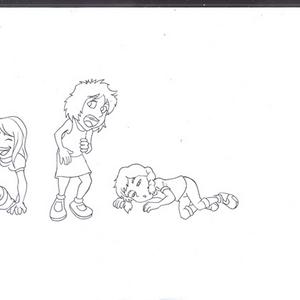}%
\includegraphics[width=0.11\linewidth]{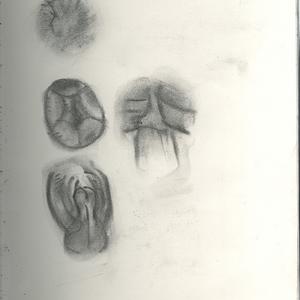}%
\\
\tabletitle{oilpaint} &
\includegraphics[width=0.11\linewidth]{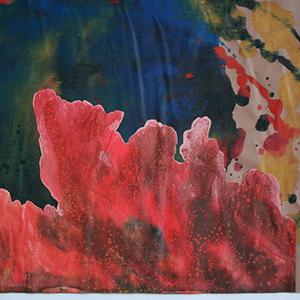}%
\includegraphics[width=0.11\linewidth]{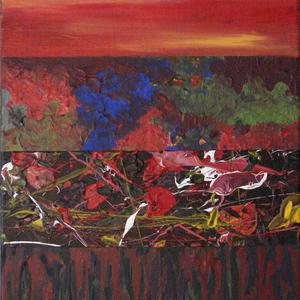}%
\includegraphics[width=0.11\linewidth]{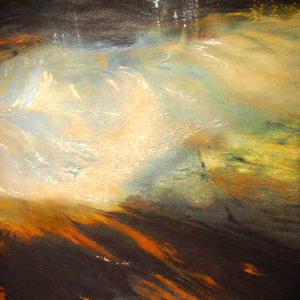}%
\includegraphics[width=0.11\linewidth]{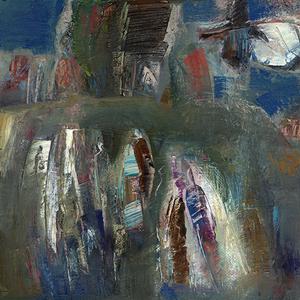}%
\includegraphics[width=0.11\linewidth]{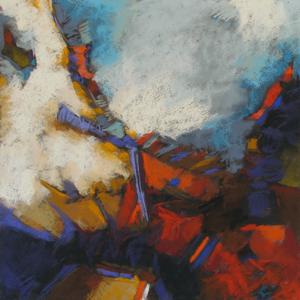}%
\includegraphics[width=0.11\linewidth]{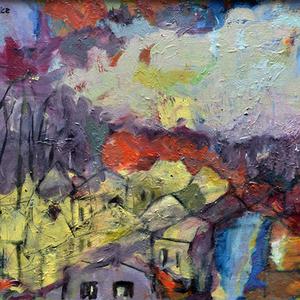}%
\includegraphics[width=0.11\linewidth]{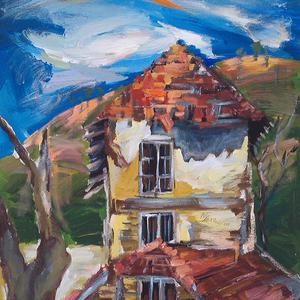}%
\includegraphics[width=0.11\linewidth]{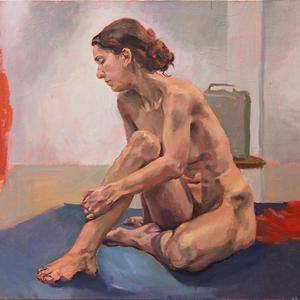}%
\\
\tabletitle{pen-ink} &
\includegraphics[width=0.11\linewidth]{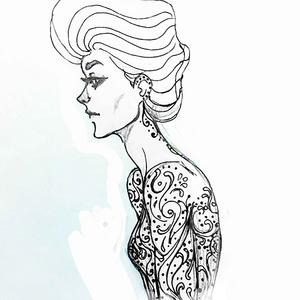}%
\includegraphics[width=0.11\linewidth]{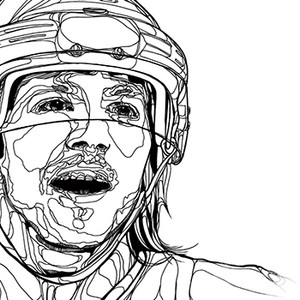}%
\includegraphics[width=0.11\linewidth]{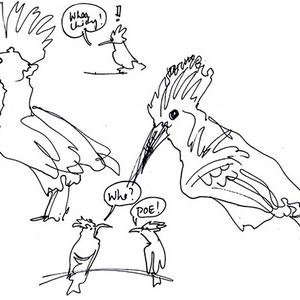}%
\includegraphics[width=0.11\linewidth]{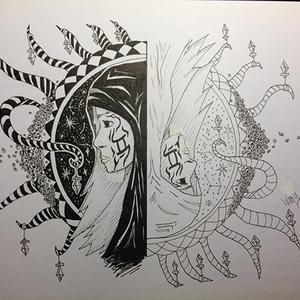}%
\includegraphics[width=0.11\linewidth]{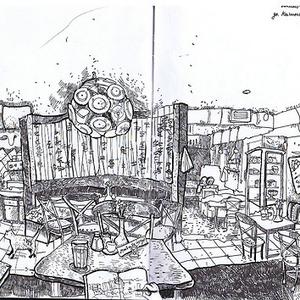}%
\includegraphics[width=0.11\linewidth]{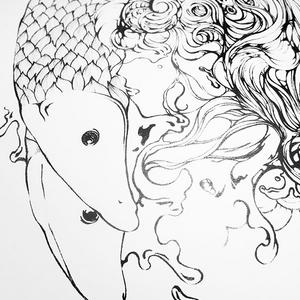}%
\includegraphics[width=0.11\linewidth]{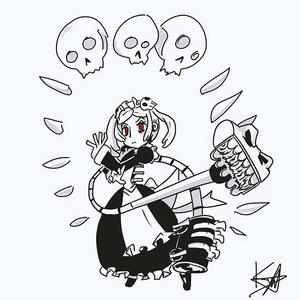}%
\includegraphics[width=0.11\linewidth]{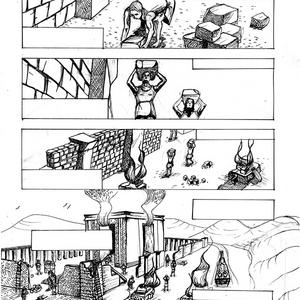}%
\\
\tabletitle{vectorart} &
\includegraphics[width=0.11\linewidth]{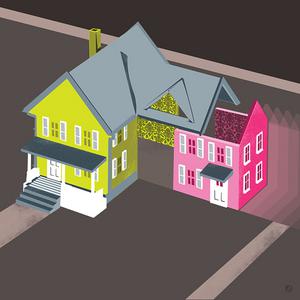}%
\includegraphics[width=0.11\linewidth]{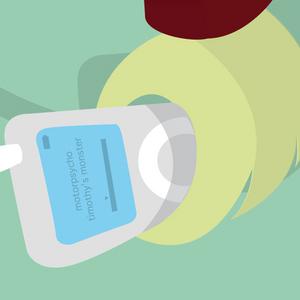}%
\includegraphics[width=0.11\linewidth]{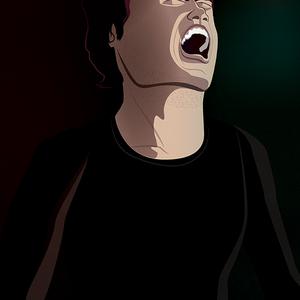}%
\includegraphics[width=0.11\linewidth]{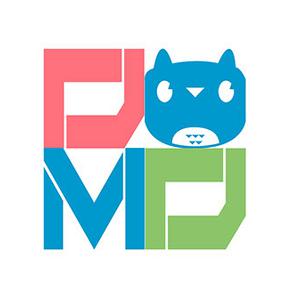}%
\includegraphics[width=0.11\linewidth]{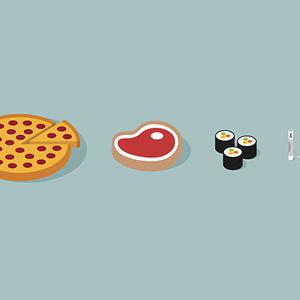}%
\includegraphics[width=0.11\linewidth]{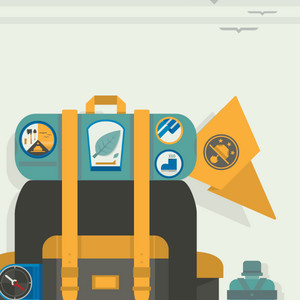}%
\includegraphics[width=0.11\linewidth]{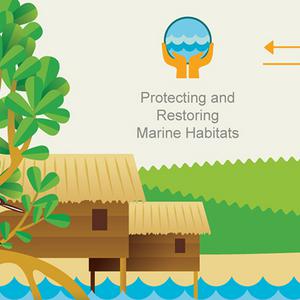}%
\includegraphics[width=0.11\linewidth]{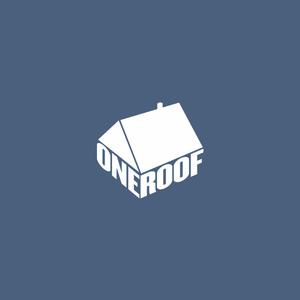}%
\\
\tabletitle{watercolor} &
\includegraphics[width=0.11\linewidth]{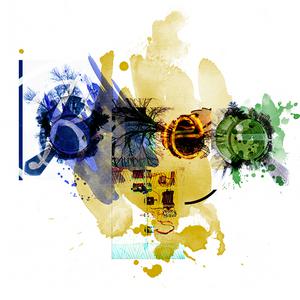}%
\includegraphics[width=0.11\linewidth]{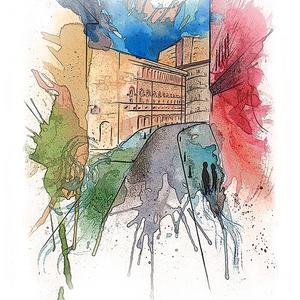}%
\includegraphics[width=0.11\linewidth]{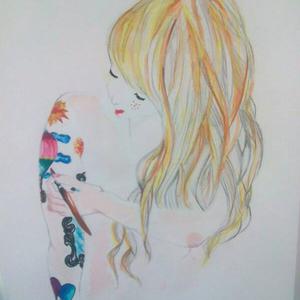}%
\includegraphics[width=0.11\linewidth]{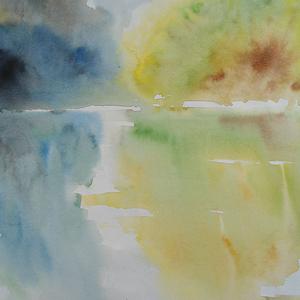}%
\includegraphics[width=0.11\linewidth]{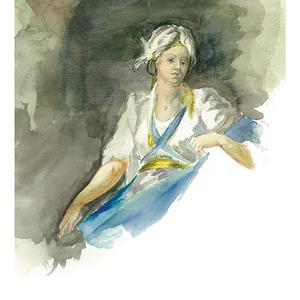}%
\includegraphics[width=0.11\linewidth]{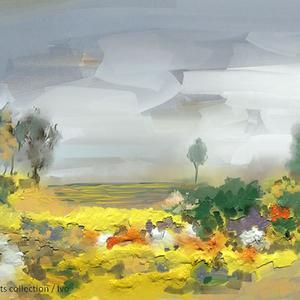}%
\includegraphics[width=0.11\linewidth]{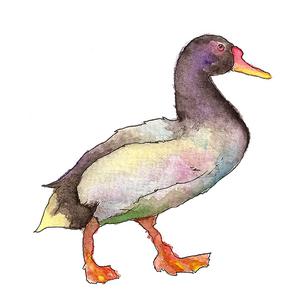}%
\includegraphics[width=0.11\linewidth]{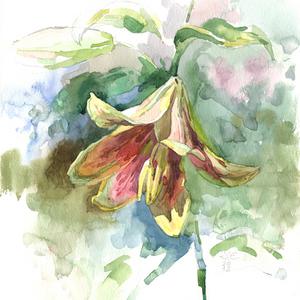}%
  \end{tabular}
}
\caption{\label{fig:example-images}Example images from Behance Artistic Media. We encourage the reader to zoom in for more detail.}
\end{figure}

Artwork on Behance spans many fields, such as sculpture, painting, photography, graphic design, graffiti, illustration, and advertising. Graphic design and advertising make up roughly one third of Behance. Photography, drawings, and illustrations make up roughly another third. This artwork is posted by professional artists to show off samples of their best work. We encourage the reader to visit \url{http://behance.net} to get a sense of the diversity and quality of imagery on this site. Example images from Behance are shown in Fig.~\ref{fig:example-images}.

\textbf{Selecting attribute categories.} In this work, we choose to annotate our own artistic binary attributes. Attribute names are rendered in \textsf{sans serif font}. Our attributes capture three categorization facets:
\begin{itemize}[topsep=0pt,itemsep=-1ex,partopsep=1ex,parsep=1ex]
\item \textbf{Media attributes:} We label images created in \textsf{3D computer graphics}, \textsf{comics}, \textsf{oil painting}, \textsf{pen ink}, \textsf{pencil sketches}, \textsf{vector art}, and \textsf{watercolor}.
\item \textbf{Emotion attributes:} We label images that are likely to make the viewer feel \textsf{calm/peaceful}, \textsf{happy/cheerful}, \textsf{sad/gloomy}, and \textsf{scary/fearful}.
\item \textbf{Entry-level object category attributes:} We label images containing \textsf{bicycles}, \textsf{birds}, \textsf{buildings}, \textsf{cars}, \textsf{cats}, \textsf{dogs}, \textsf{flowers}, \textsf{people}, and \textsf{trees}.
\end{itemize}
%\aaron{perhaps highlight attribute names with different typesetting, like \textsf{happy/cheerful}, \textsf{pen-and-ink}. As it is, it's hard to tell that these are distinct labels. The list could also be made more explicity. "Content attributes. We label 9 object classes: dogs, people, buildings, ...}
%\aaron{alphabetize the lists? are they in a particular order for a reason?}

We chose these attributes as follows:
The seven media attributes were chosen on the expert advice of a resident artist to roughly correspond with the genres of artwork available in Behance that are easy to visually distinguish. Our goal is to strike a balance between distinctive media while covering the broad range available in Behance. For instance, oil paint and acrylic are considered to be different media by the artistic community, but are very hard for the average crowdworker to distinguish visually.
%\todo Mention: We aimed for visually distinct attributes. Oil and acrylic are different media, but very hard to distinguish visually (if not impossible).
The four emotion attributes are seen on Plutchik's \emph{Wheel of Emotions}~\cite{Plutchik2001}, a well-accepted model for emotions that was also used in~\cite{Jou2015}. From this model, we chose the emotions that are likely to be visually distinctive.
The content attributes represent entry-level object categories and were chosen to have some overlap with Pascal VOC while being representative of Behance content. We focus on entry-level categories because these categories are likely to be rendered in a broad range of styles throughout Behance.%these are the easiest to identify across different styles of artwork. %More fine-grained distinctions between plant species or dog breeds may be possible for some images, but artists may not be aware of these differences as they render their art; we leave this problem to future work.
% \todo I haven't explained what I'm doing with the dataset yet. This paragraph should be moved into some kind of experiment that comes after we show how we label the dataset.

Although this work is only concerned with a small set of labels (arguably a proof-of-concept), the dataset we release could itself be the basis for a real PASCAL/COCO-sized labeling effort which requires consortium-level funding.

\textbf{Tags are noisy.} Behance contains user-supplied tags, and one may wonder whether it is feasible to train attribute classifiers directly from these noisy tags alone, such as in previous work~\cite{Izadinia2015,MisraNoisy16}.
However, unlike that work, we cannot create our dataset from tags alone for two reasons. First, not all of our attributes have corresponding tags. Second, tags are applied to each \emph{project}, not each image. For example, even though a project called ``Animal sketches 2012'' may have the ``Dog'' tag, we do not know which image that tag should apply to. Training on tags alone is too noisy and reduces the final classifier precision. To demonstrate, we train a binary classifier on the ``Cat'' tag, but from manual inspection, it only learns to distinguish different small animals and is not fine-grained enough to find cats; see Fig.~\ref{fig:tag-only-comparison}. The precision of cats among the top 100 detections is only about 36\%. To increase this accuracy, we must rely on human expertise to collect labels.

%- Chen’s biggest worry: Why did we pick the object categories we did? Entry-level attributes are rendered most robustly in artwork; VOC overlap. Media categories: These are the genres recommended by an art expert. Emotion categories: Four spokes of the “Wheel of emotion” Must have good justification!
%- Behance doesn't aim to be a historically accurate archive of artwork. It's completely contemporary. To avoid reviewers' comments about being accurate to artistic styles, we want to explain how Behance represents a sampling of styles that occur today.
%The primary place where professional and commercial artists share their portfolios.

%Our attributes; how we chose them ...

\subsection{Annotation pipeline}\label{sec:crowdsourcing}
%%
%% What parts are most important?
%% - Dataset statistics
%% - Quality
%% - Overall pipeline
%% - Classifier
%% - Crowd stuff
%%

%Several works~\cite{Yu2015LSUNCO,Krause2016TheUE,Cui2015FinegrainedCA} show how to use deep learning to amplify human effort. The design of our crowdsourced dataset collection process is loosely based on the LSUN dataset annotation pipeline~\cite{Yu2015LSUNCO}, which builds a very large-scale object detection dataset using a combination of deep learning and crowdsourcing.

\begin{figure}[t]
\centering
\includegraphics[width=0.9\linewidth]{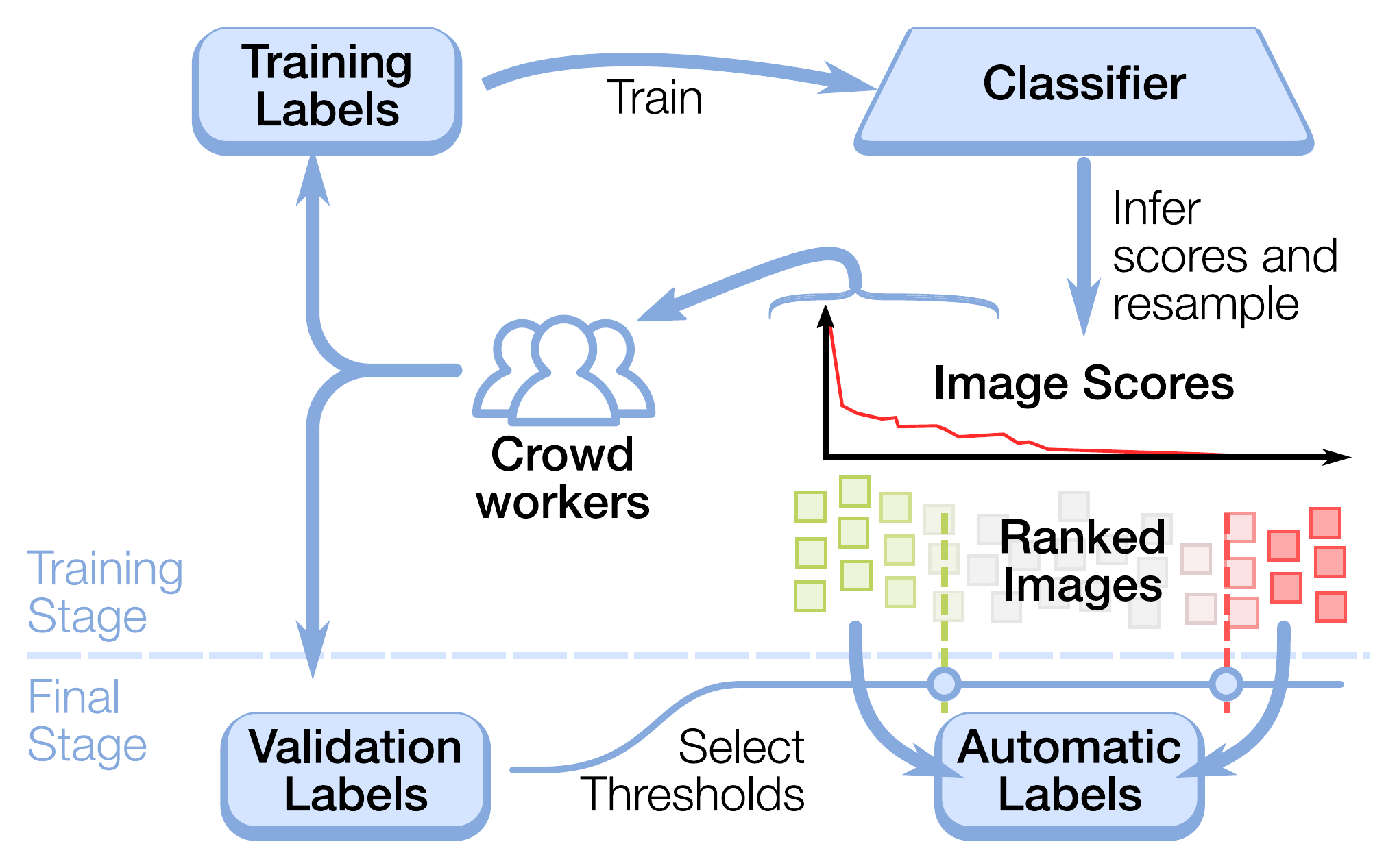}
\caption{\label{fig:crowdsourcing-process}A diagram of our crowdsourcing pipeline. First, we train a set of classifiers on all labels collected so far. We then use this classifier to rank a random sample of images. High-scoring images are sent back to the crowd, and the resulting labels are added to the training and validation set. After four iterations, the validation set is used to select positive and negative thresholds with certain precision and recall targets. Images meeting these thresholds are added to the automatic label set.
}
\end{figure}

Our dataset requires some level of human expertise to label, but it is too costly to collect labels for all images. To address this issue, we use a hybrid human-in-the-loop strategy to incrementally learn a binary classifier for each attribute.
Our hybrid annotation strategy is based on the LSUN dataset annotation pipeline described in~\cite{Yu2015LSUNCO}, which itself shares some similarity with other human-in-the-loop collection systems~\cite{Yu2015LSUNCO,Krause2016TheUE,Cui2015FinegrainedCA}. An overview of this process is shown in Fig.~\ref{fig:crowdsourcing-process}.
At each step, humans label the most informative samples in the dataset with a single binary attribute label. The resulting labels are added to each classifier's training set to improve its discrimination. The classifier then ranks more images, and the most informative images are sent to the crowd for the next iteration. After four iterations, the final classifier re-scores the entire dataset and images that surpass a certain score threshold are assumed to be positive. This final threshold is chosen to meet certain precision and recall targets on a held-out validation set. This entire process is repeated for each attribute we wish to collect.

\textbf{Crowdsourcing task.}% Due to space constraints, we cover the details abuot the actual Mechanical Turk crowdsourcing task in the supplementary material.
The heart of our human-in-the-loop system is the actual human annotation task. We collect annotations for each attribute independently.
To do this, we rely on Amazon Mechanical Turk, a crowdsourced marketplace. Crowdworkers (\emph{``Turkers''}) complete Human Intelligence Tasks for a small cash payment.
In each HIT for a given attribute, we show the Turker 10 handpicked positive/negative example images and collect 50 binary image annotations. Turkers indicate whether each image has the attribute of interest. Each HIT only collects labels for a single attribute at a time to avoid confusion. For quality control, we show each image to two separate Turkers and only use answers where both Turkers agree.
We also collect sparse text annotations for a subset of these images.
Every 10 images, we present an annotation recently provided by the Turker and ask for a brief 3-word caption to justify their choice. %For emotion attributes, we ask why the image might or might not make an average Turker feel that emotion; for media attributes, we ask how the Turker knows the image was or was not drawn with that medium; for content attributes, we merely ask what the object of interest looks like.
%The Turker must write at least three words before continuing, but captions are not checked for grammar or coherence.
This has the effect of encouraging annotators to carefully consider and justify their choices.
%Turkers found this extra captioning task to be annoying and to slow them down; however, it seemed to greatly increase the quality of the provided attribute labels based on manual inspection of results.

%These annotations also provide useful clues about what qualities workers use to describe style.
%The words that maximize TF/IDF scores among positive annotations are informative: workers tend to use nouns when describing object categories (such as ``bouquet'', ``rose'', ``petal'', ``vase'' for \textsf{Flower}) and visual adjectives when describing media and emotion (such as ``translucent'', ``frayed'', ``blotchy'', ``bleeding'', ``overlap'' for \textsf{Watercolor}).
%The supplementary material contains more examples of informative words. These annotations will be released alongside the final dataset.

% \aaron{we could break out the annotations into a separate paragraph, because they might be a significant part of the dataset. Here we'd say that we release these annotations with the dataset, and give examples of them here, and add more analysis of them in the next section. e.g., topic analysis of which annotations are discriminative for each attribute.}

%Annotators were paid \$0.30 for each HIT task. The median task completion time was 251 seconds, leading to an average hourly pay of \$4.30, but the fastest 25\% of workers earned \$6.40 for their efforts.

It is always important to balance the trade-off between squeezing high-quality work out of annotators while being respectful of their effort and abilities. The subjectivity of our task makes this trade-off harder to manage. %For this reason, we wished to minimize the amount of automated quality checking.
%%Beyond occasionally asking Turkers for justification, we did not feel a need to force them to surpass accuracy thresholds on ``gold standard'' tasks.
To address this issue, we prevent spam by only accepting work from crowdworkers who previously completed 10,000 MTurk tasks with a 95\% acceptance rate, and we only use labels where both workers agree.
Finally, the quality of the entire dataset is ensured by setting appropriate label thresholds on held-out validation data; see Sec.~\ref{sec:quality}
%We did not reject any HIT responses and annotators were always paid for their time, opting to instead ignore low-quality responses without consensus. From manual inspection, we found that with adequate examples, our annotators generally understood the task and answered to the best of their abilities.
%\aaron{this requires some justification. how do we know we have good labels? this procedure doesn't sound robust to spammers or poor workers.}

\textbf{Iterative learning.} Starting from a small handpicked initial label set, the dataset is enlarged by an iterative process that alternates between training a classifier on the current label set, applying it to unlabeled images, and sending unconfident images back to the crowd for more labeling.
%
%We start the process by constructing a \emph{tentative training set}. For media and content attributes, we sample images with handpicked tags as positives and random images as negatives. For example, the \textsf{Dog} attribute is seeded by a classifier trained on positive images from Behance tagged with ``Dog''. The first classifier is trained on this tentative training set with the expectation that this classifier's guesses will be quickly refined by the crowd.
%For emotion images, there are no suitable tags, so we start by collecting a training set from the crowd, randomly sampled from photography and fine art fields.
%
On each iteration, we train a deep learning classifier using 10/11$^{\text{ths}}$ of the total collected crowd labels. The last 1/11$^{\text{th}}$ is always held out for validation. We apply this classifier to the entire dataset.
The crowd then labels 5,000 images that score higher than a threshold set at 50\% precision measured on validation data. This way, we show our Turkers a balance of likely-positive and likely-negative images each time.
%To select the next round of images to show to the crowd, we define an ``interestingness'' score threshold on the validation set such that the precision of relevant validation images above this threshold is 50\%. The crowd labels 5,000 ``interesting'' images above this threshold. This way, the crowd always sees an even split of likely-positive and likely-negative images. The resulting crowd labels are added to the training set for the next iteration.

After four iterations, we arrive at a final classifier that has good discrimination performance on this attribute. We score the entire dataset with this classifier and use thresholds to select the final set of positives and negatives. The positive score threshold is chosen on validation data such that the precision of higher-scoring validation images is 90\%, and the negative threshold is chosen such that the recall of validation images above this threshold is 95\%. In this way, we can ensure that our final labeling meets strict quality guarantees.

% The hybrid loop
% - For content and artistic media attributes, we first collect a tentative positive set used to train a first classifier from tags. For emotion attributes, we show the crowd randomly sampled images from the photography and fine art fields.

It is important to note that the resulting size of the dataset is determined solely by the number of relevant images in Behance, our desired quality guarantees, and the accuracy of the final classifier. A better attribute classifier can add more images to the positive set while maintaining the precision threshold. If we need more positive data for an attribute, we can sacrifice precision for a larger and noisier positive set. %{\color{red} \todo consider put some threshold-dataset size pairs here or in 3.2.}

\textbf{Classifier.} For content attributes, our classifier is a fine-tuned 50-layer ResNet~\cite{DBLP:journals/corr/HeZRS15} originally trained on ImageNet. For emotion and media attributes, we found it better to start from \emph{StyleNet}~\cite{Fang2015CollaborativeFL}. This model is a GoogLeNet~\cite{DBLP:journals/corr/SzegedyLJSRAEVR14}, fine-tuned on a style prediction task inferred from user behavior. Each network is modified to use binary class-entropy loss to output a single attribute score. To avoid overfitting, we only fine-tune for three epochs on each iteration. See Fig.~\ref{fig:example-images} for examples of Behance images.

\begin{figure}
\centering
  \includegraphics[width=0.9\linewidth]{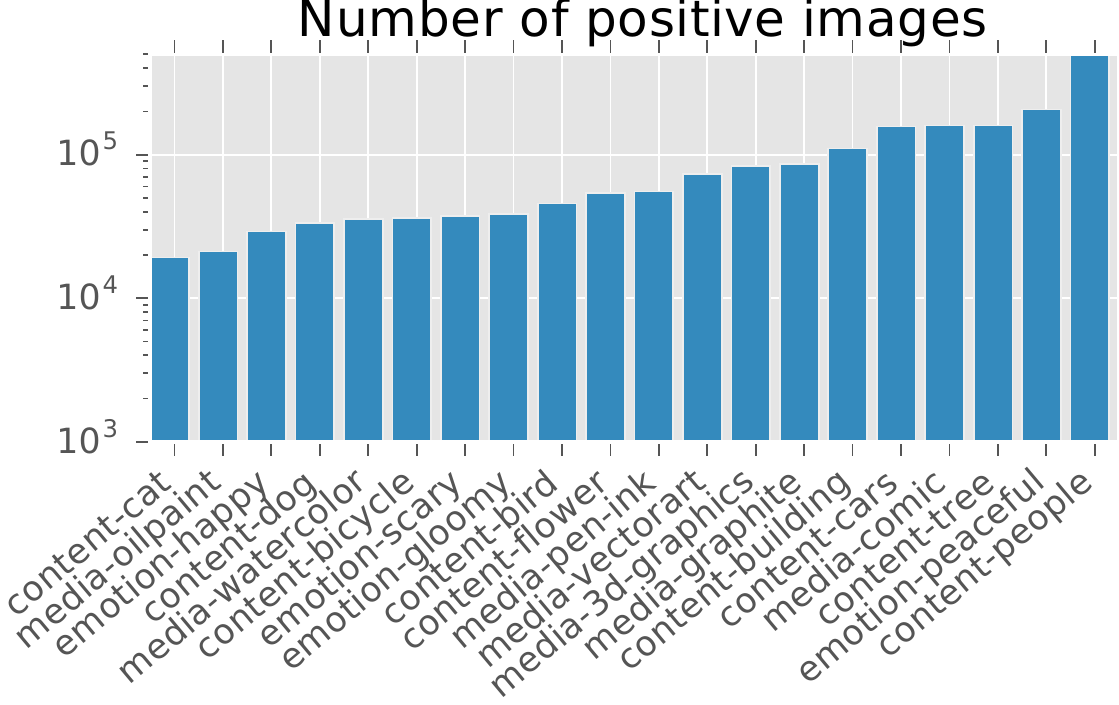}
  \caption{\label{fig:size} Top: Number of positive images in the final set.} %Bottom: Amount of amplification for each attribute (number of automatic labels divided by number of crowd labels)}
\end{figure}

\textbf{Resulting dataset statistics}
Our final dataset includes positive and negative examples for 20 attributes. The median number of positive images across each attribute is 54,000, and the median number of negative images is 8.7 million. The ``People'' attribute has the most positive images (1.74 million). Humans are commonly featured as art subjects, so this is not surprising. The attribute with the least positives is ``Cat'' with 19,244 images. We suspect this is because our final labeling model cannot easily distinguish cats from other cat-like renditions. Cats on Behance are commonly rendered in many different styles with very high intra-class variation. Statistics for all attributes are shown in Fig.~\ref{fig:size}.

Our automatic labeling model can amplify the crowd's annotation effort. The ratio of automatic positive labels to crowd-annotated positive labels is 17.4. The amplification factor for negative labels is much higher---about 505---because automatic systems can quickly throw away easy negatives to focus the crowd's attention on potentially relevant images.
%505 averaged across all attributes. Here, ``amplification'' is defined as the number of automatically inferred labels divided by the number of images seen by the crowd. When labeling a dataset as large and diverse as Behance, automatic systems can quickly throw away easy negatives, focusing the crowd's attention on potentially relevant images. This means most of the amplification effect comes from negative images. If we alternatively define amplification as the number of automatically-labeled \emph{positive} images divided by the number of crowd-labeled positive images, the average amplification factor is 17.4.

% \subsection{\todo Show dataset modes (E1)}

\begin{figure}[t]
  \centering
  \includegraphics[width=0.9\linewidth]{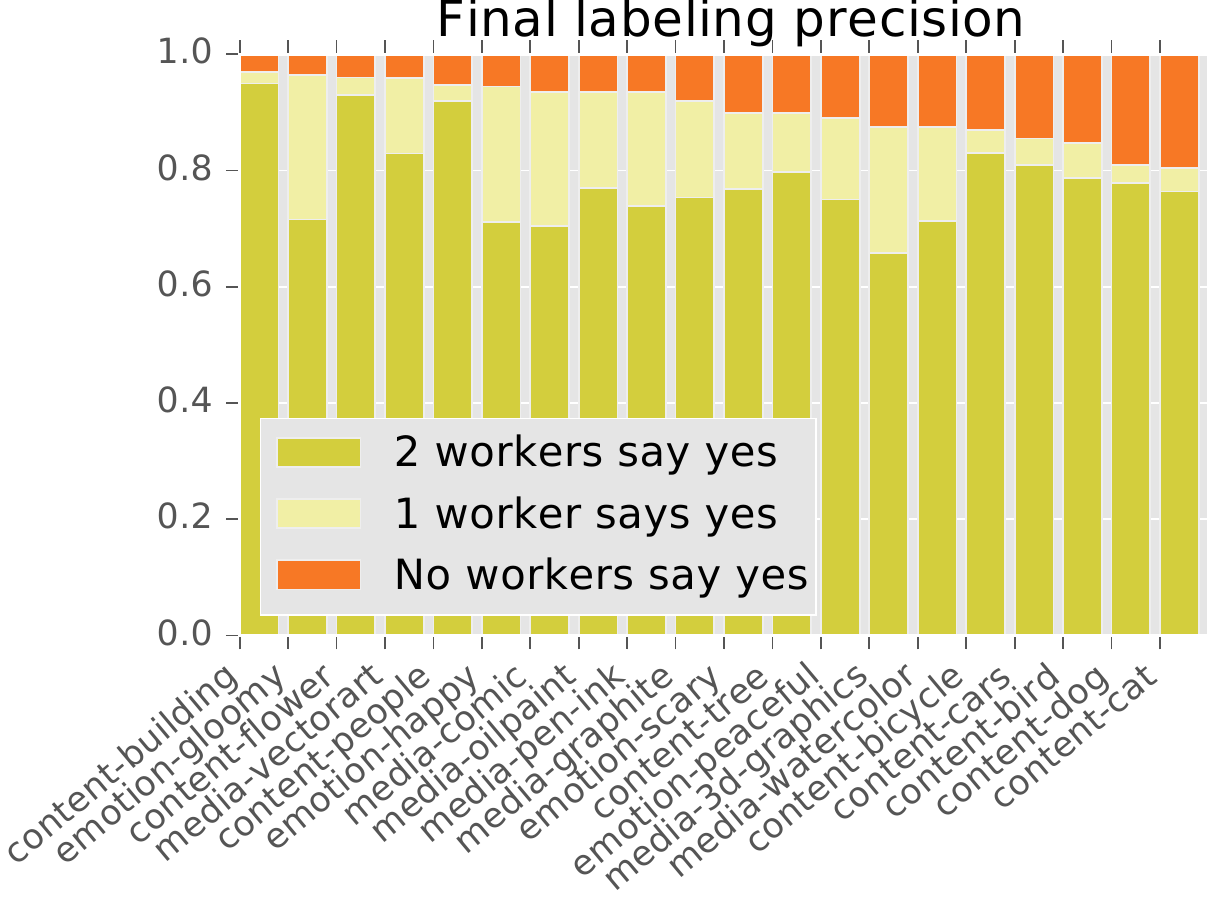}
  \caption{\label{fig:label-quality} Final quality assurance: Showing worker agreement of automatically-labeled positive images in the final dataset.}
\end{figure}

\textbf{Final quality assurance}
\label{sec:quality}
As a quality check, we tested whether the final labeling set meets our desired quality target of 90\% precision. For each attribute, we show annotators 100 images from the final automatically-labeled positive set and 100 images from the final negative set using the same interface used to collect the dataset. Fig.~\ref{fig:label-quality} shows worker agreement on the positive set as a proxy for precision. The mean precision across all attributes is 90.4\%, where precision is the number of positive images where at least one annotator indicates the image should be positive.
%Across all attributes, at least one worker indicated the image was negative for 98.9\% images in the negative set, surpassing our original recall target of 95\%.
These checks are in addition to our MTurk quality checks: we only use human labels where two workers agree and we only accept work from turkers with a high reputation who have completed 10,000 tasks at 95\% acceptance.

\section{Experiments}
We can use Behance Artistic Media to study recognition across artistic domains as well as aesthetics and style. First, we investigate the \emph{representation gap} between objects that appear in everyday photographs and objects that appear in artwork. We find that ordinary object detectors do not adequately recognize artistic depictions of objects, showing that there is room for improvement.
The existence of this gap leads us to explore the relationship between object representations as rendered across different artistic media. We pose this as a \emph{domain transfer} problem and measure the extent to which knowledge about objects in one medium can apply to objects in an unseen medium.
In addition to objects, we briefly consider \emph{style and aesthetics} by comparing different features on emotion/media classification and using our style labels to improve aesthetic prediction tasks on other art datasets.
Finally, we conclude with an experiment of \emph{learning feature spaces (feature disentangling)} to build a task-specific search engine that can search for images according to their content, emotion, or media similarity.

\begin{figure}[t]
  \centering
\includegraphics[width=0.4\linewidth]{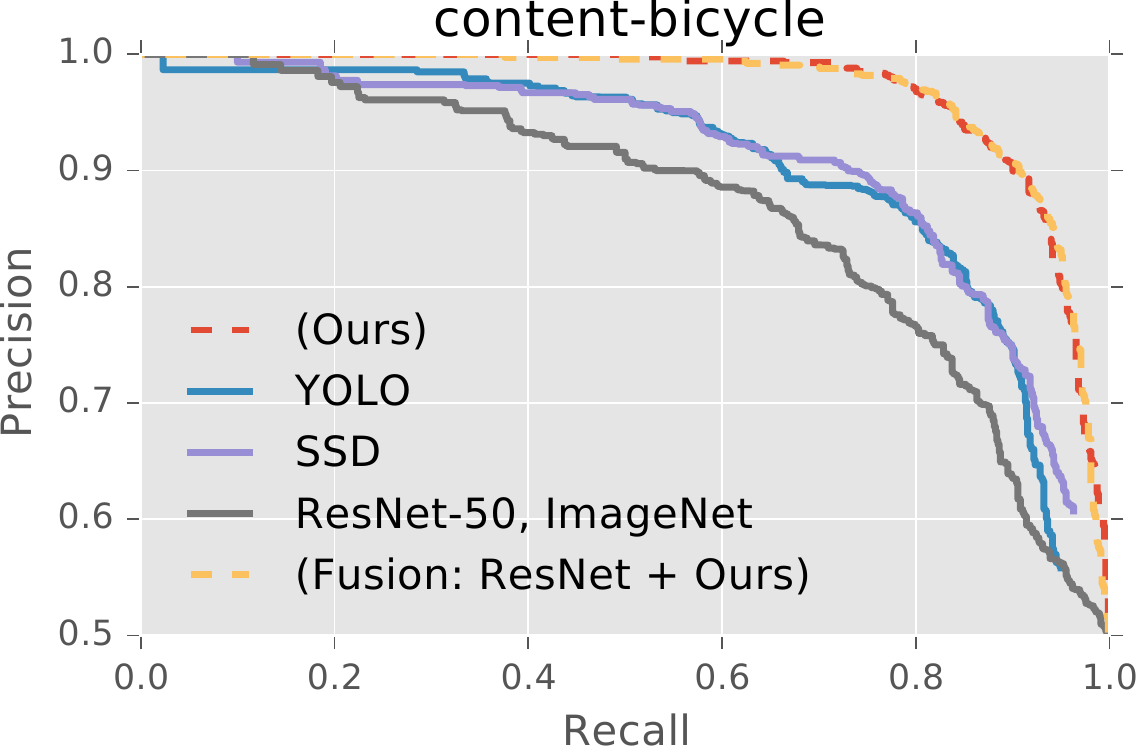}
\includegraphics[width=0.4\linewidth]{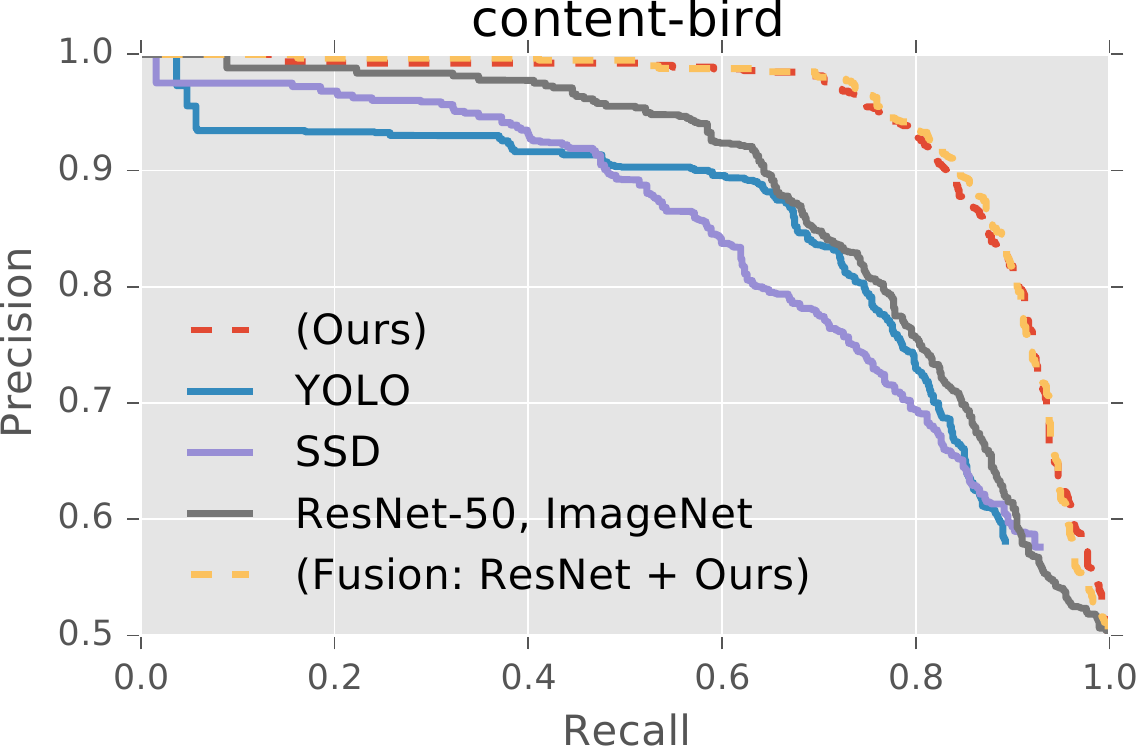}
\\
\includegraphics[width=0.4\linewidth]{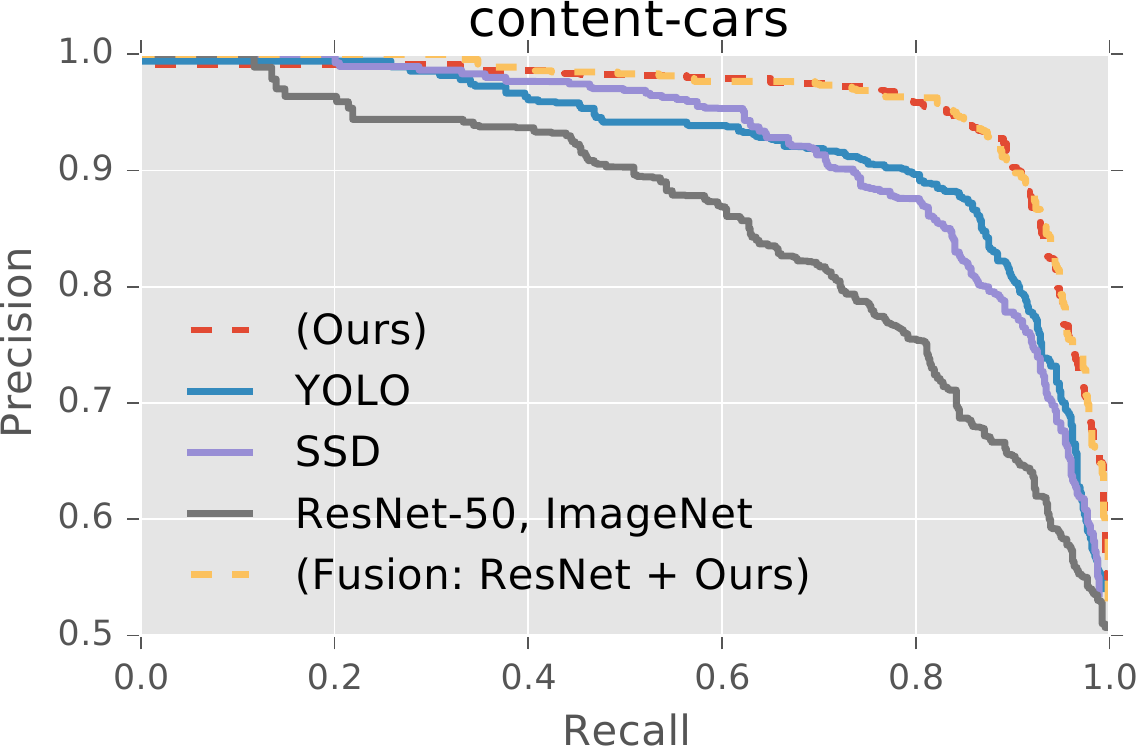}
\includegraphics[width=0.4\linewidth]{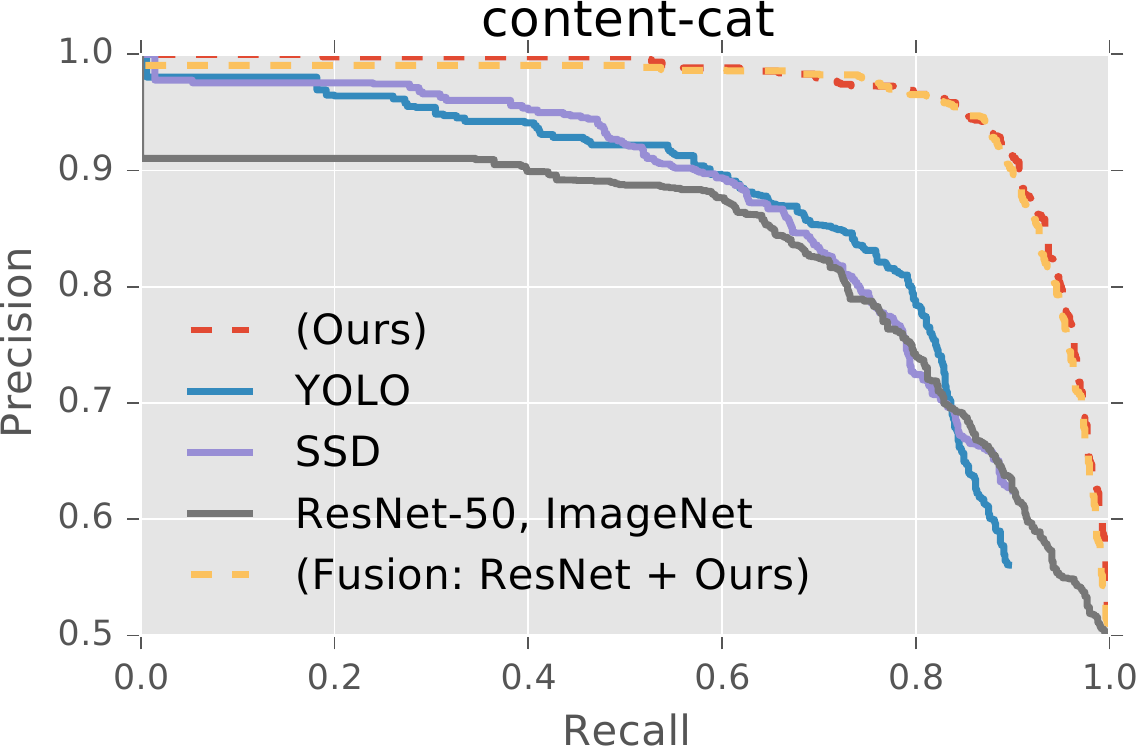}
\\
\includegraphics[width=0.4\linewidth]{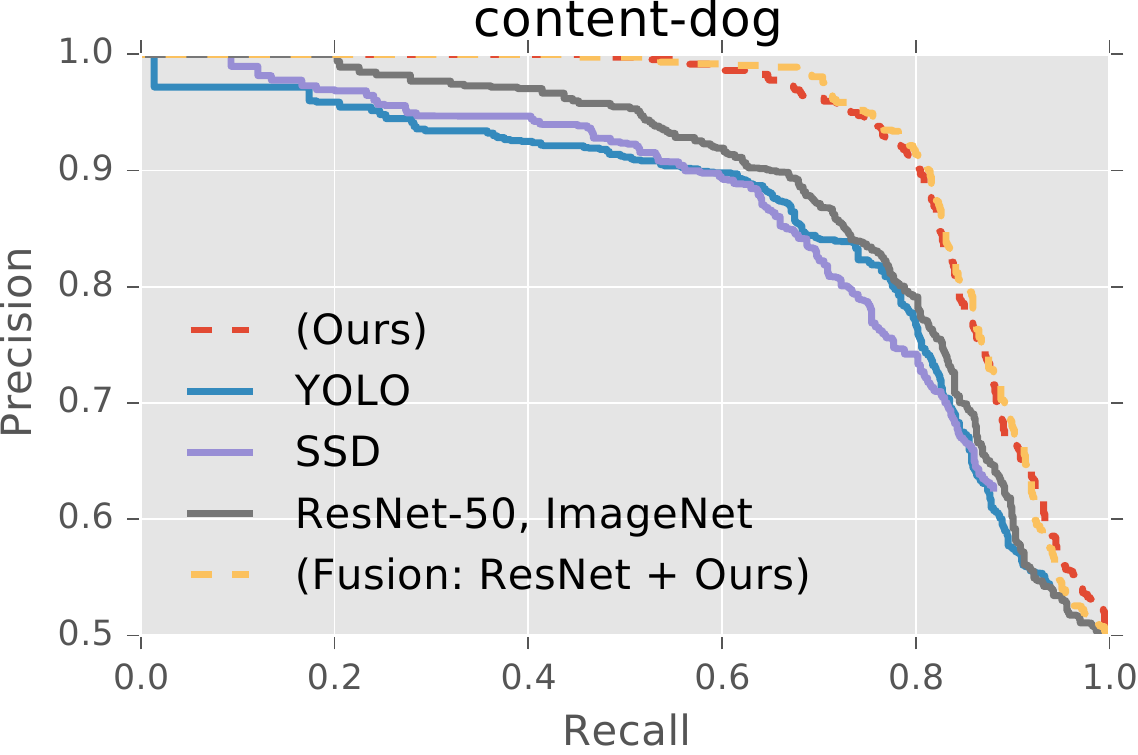}
\includegraphics[width=0.4\linewidth]{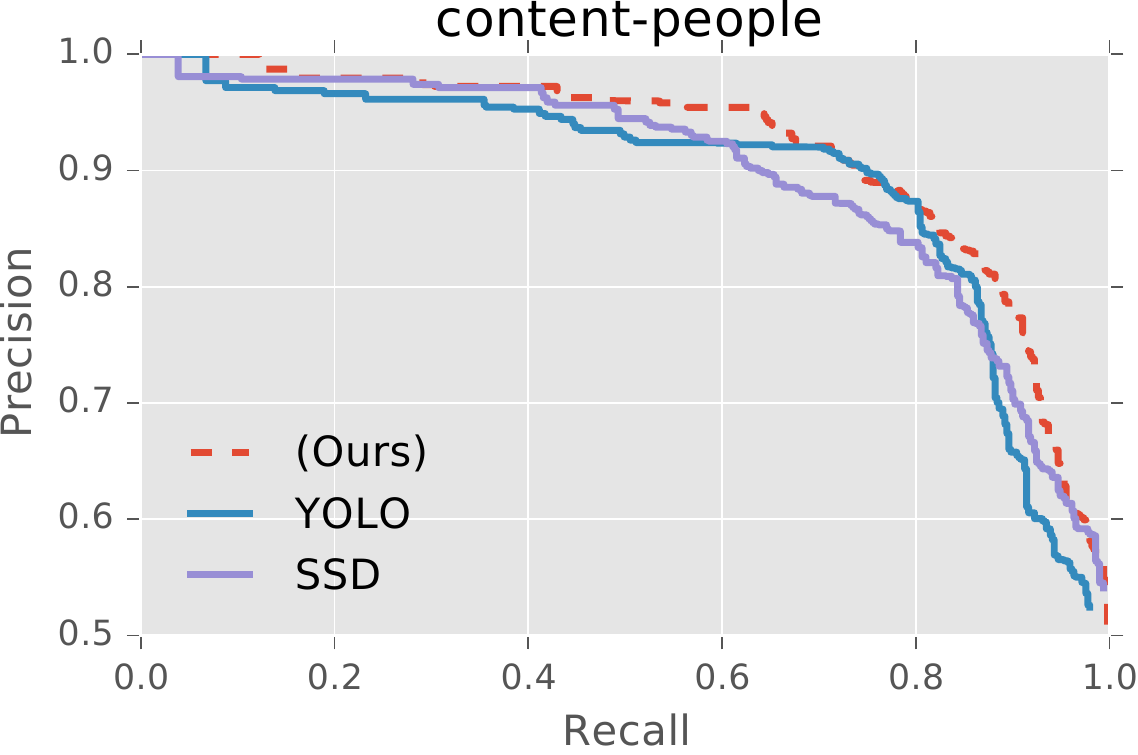}
\caption{\label{fig:content-objdetector-pr} PR curves for object categories comparing our model, YOLO, SSD, ResNet-50, and fusion of ours and ResNet-50.}
\end{figure}

\subsection{Bridging the representation gap}
\textbf{Detecting objects in artwork.}
How different are objects in everyday photographs compared to the stylized objects found in our dataset? We expect that existing pre-trained object detectors might not recognize objects in artwork because existing object detectors trained on ImageNet or VOC are only exposed to a very narrow breadth of object representations. Objects in photographs are constrained by their real-world appearance.

To investigate the representation gap between our dataset and everyday photographs,
we consider 6 content attributes that correspond to Pascal VOC categories: \textsf{Bicycle}, \textsf{Bird}, \textsf{Cars}, \textsf{Cat}, \textsf{Dog}, \textsf{People}. We then extract scores for these attributes using two object detectors trained on VOC: YOLO~\cite{Redmon2016} and SSD~\cite{Liu2015}. For the sake of comparison, we use these detectors as binary object classifiers by using the object of interest's highest-scoring region from the detector output. We also compare to ResNet-50 classifiers~\cite{DBLP:journals/corr/HeZRS15} trained on ImageNet, taking the maximum dimension of the ImageNet synsets that correspond with the category of interest. In this way, we can measure how well existing object detectors and classifiers already find objects in art without extra training. We also compare to our final attribute classifier trained in Sec.~\ref{sec:crowdsourcing}, the fine-tuned ResNet-50 that was used to automatically label the final dataset.% {\color{red} explain a bit more to recap readers what is trained in 3.1}

%- In Section 4.1 (L688), the authors say that they used 5000 positive and negative human labeled images for each attribute for validation. It is not clear how these images were obtained given that the system underwent 4 iterations of 5000 human labels each where only 10% of the images were saved for validation i.e., 2000 images total (assume 1000 positive and 1000 negative). If human labeled images that were used for training the deep network were used for validation, the results are clearly invalid. If not, what procedure was used to obtain this validation set?
We evaluate these methods on 1,000 positives and 1,000 negatives on each attribute's \emph{human-labeled} validation set to avoid potential bias from the automatic labeler. The results are shown as precision/recall curves in Fig.~\ref{fig:content-objdetector-pr} and AP is shown in Tab.~\ref{tab:results}. Vision systems trained on photography datasets like VOC (YOLO, SSD) and ImageNet (RN50) perform worse than vision systems that saw objects in artwork during training. From manual inspection, most false negatives of these systems involve objects rendered with unique artistic styles. Specific failure cases are shown in Fig.~\ref{fig:content-style}.
We can improve performance slightly by fusing ImageNet and Behance scores together with a simple linear combination. The resulting ``Fusion'' model performs slightly better than our own model and ResNet-50 on all but two attributes.
These results show that in terms of object recognition, there is a \emph{representational gap} between photography and artwork.

\begin{table}[t]
  \scalebox{0.95}{
  \begin{tabular}{r|cccc|c}
    AP & Ours & Yolo & SSD & RN50 & Fusion \\
    \hline
    \textsf{Bicycle} & \textbf{0.9703} & 0.9008 & 0.9116 & 0.8702 & 0.9704 \\
    \textsf{People} & \textbf{0.9103} & 0.8863 & 0.8952 & ---$^1$ & --- \\
    \textsf{Bird} & \textbf{0.9400} & 0.8516 & 0.8387 & 0.8768 & 0.9453 \\
    \textsf{Cat} & \textbf{0.9660} & 0.8583 & 0.8620 & 0.8026 & 0.9501 \\
    \textsf{Cars} & \textbf{0.9551} & 0.9140 & 0.9194 & 0.8519 & 0.9628 \\
    \textsf{Dog} & \textbf{0.9272} & 0.8510 & 0.8582 & 0.8818 & 0.9293 \\
    \hline
    Average & \textbf{0.9448} & 0.8770 & 0.8801 & 0.8567 & 0.9512
  \end{tabular}
  }

  \vspace{0.5ex}
  \caption{\label{tab:results} Average precision across different VOC categories using our model, YOLO, SSD, ResNet-50, and fusion of ours and ResNet-50. $^1$: We do not report \textsf{people} results because there are relatively few ImageNet \textsf{people} categories.}
\end{table}

\textbf{Object representation across artistic media.}
The existence of this representational gap leads us to question how objects are represented across different artistic media.
How well do models trained on one medium generalize to unseen media, and which media are most similar?
We can answer these questions within the context of domain adaptation, which has been extensively studied in the vision literature~\cite{Csurka2017DomainAF,7078994}.
A good model should know that although cats rendered in drawings are more ``cartoony'' and abstract than the realistic cats seen in oil paint and ImageNet, they both contain the same ``cat'' semantic concept, even though the context may vary.

We retrieve the 15,000 images that maximize $\sigma(x_{i,c})\sigma(x_{i,m})$ for every pair of content and media labels $(c,m)$, where $\sigma$ is the sigmoid function and $x_i$ is that image's label confidence scores. We set aside 1/11th of these as the validation set. Note that this validation set is a strict subset of the validation set used to train the automatic labeler. We then fine-tune a pre-trained ResNet for one epoch. The last layer is a 9-way softmax.

\begin{figure}[t]
\includegraphics[width=\linewidth]{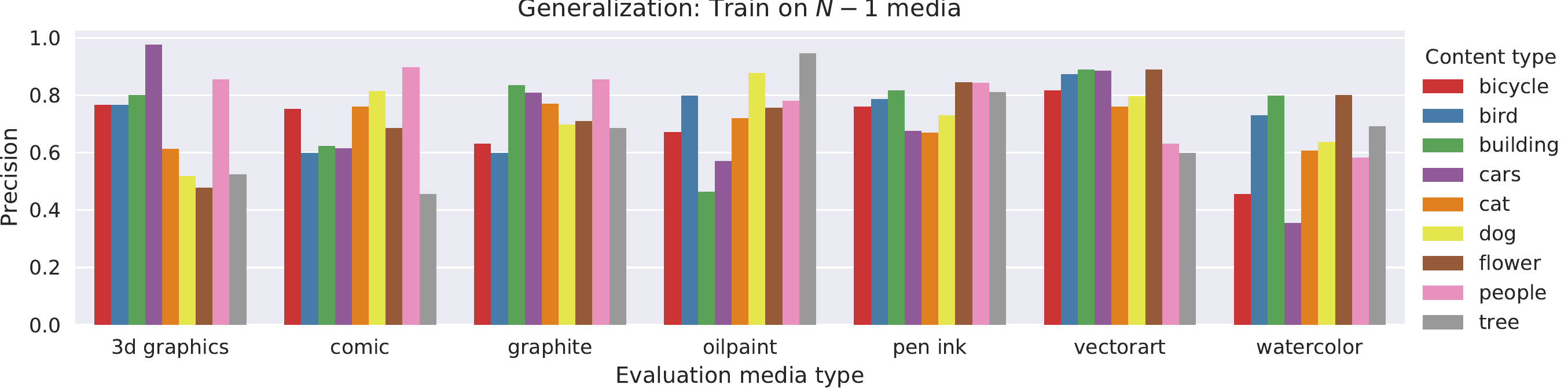}
\caption{\label{fig:generalization-media}Object recognition performance on an unseen domain. We evaluate an object classifier on the given artistic medium while training it on the 6 other media.}
\end{figure}

In the first set of experiments, we measure an object classifier's ability to generalize to an unseen domain by learning the representation styles across the other 6 media types and evaluating on only the 7th media type.
Results are summarized on the last row of Tab.~\ref{tab:gen-mtx} and broken down by object categories in Fig.~\ref{fig:generalization-media}. Generally, objects that are iconic and easily recognizable within each medium have the highest performance (for example, \textsf{3D}+\textsf{cars}, \textsf{watercolor}+\textsf{flowers}, \textsf{graphite}+\textsf{people}),  but objects that are unlikely to be drawn consistently within each style have the worst generalization performance (\textsf{watercolor} \textsf{cars}/\textsf{bicycles}, \textsf{3D flowers}). Even though the frequency was controlled by sampling a constant number of images for every $(object,medium)$ pair, this could be because the artist is less familiar with uncommon objects in their medium and has more individual leeway in their portrayal choices.
%Even though the frequency was controlled by sampling a constant number of images for every $(object,medium)$ pair, this could be because the artist's familiarity with objects that are characteristic in their media constraints their portrayal choices.

These experiments reveal how well classifiers can generalize to unseen domains, but they do not reveal the correlations in object style between different media types. To capture this, our second set of experiments trains an object classifier on only a single media type and evaluates performance on a second media type.
As an additional photography medium, we also retrieve 15,000 images for each object from its corresponding ImageNet synset.
Average object classification accuracy is shown in Tab.~\ref{tab:gen-mtx}. The $N-1$ baseline model is trained on all other types. This metric gives a rudimentary comparison of the similarity between artistic media; for instance, \textsf{comic}, \textsf{graphite}, and \textsf{pen ink} are similar to each other, as are \textsf{oil paint} and \textsf{watercolor}. In addition to the gap between ImageNet and Behance (compare last two rows), these results illustrate the gap between each meium's stylistic depictions.
Our dataset can be used to explore these relationships and other similar domain adaptation problems.
%Interestingly, training on Behance and testing on ImageNet (last column) has higher performance than training on ImageNet and testing on Behance (last row). This could indicate that objects in Behance are represented more broadly than the photographs seen in ImageNet.

\newcolumntype{R}[2]{%
    >{\adjustbox{angle=#1,lap=\width-(#2)}\bgroup}%
    l%
    <{\egroup}%
}
\newcommand{\rot}[1]{\multicolumn{1}{R{90}{-0.7em}}{\hspace{-0.5em}#1}}% no optional argument here, please!

\begin{table}[t]
  \centering
\scalebox{0.7}{
%\setlength{\tabcolsep}{0.2em}
%\hspace{-2em}
%\setlength{\extrarowheight}{-14pt}
%\renewcommand{\arraystretch}{0.8}
  \begin{tabular}{rrrrrrrrr}
\diaghead(5,-2){Train Test}{\textbf{Train}}{\textbf{Test}}&\rot{3d}&\rot{comic}&\rot{graphite}&\rot{oil paint}&\rot{pen ink}&\rot{vectorart}&\rot{watercolor}&\rot{ImageNet}\\
3D &\cellcolor[rgb]{0.94,0.98,0.96} 81 & \cellcolor[rgb]{0.45,0.82,0.63} 37 & \cellcolor[rgb]{0.64,0.88,0.76} 54 & \cellcolor[rgb]{0.67,0.89,0.78} 56 & \cellcolor[rgb]{0.54,0.85,0.69} 45 & \cellcolor[rgb]{0.70,0.90,0.80} 59 & \cellcolor[rgb]{0.54,0.85,0.70} 45 & \cellcolor[rgb]{0.78,0.93,0.85} 66\\
Comic &\cellcolor[rgb]{0.78,0.93,0.85} 66 & \cellcolor[rgb]{0.90,0.97,0.93} 77 & \cellcolor[rgb]{0.79,0.93,0.86} 67 & \cellcolor[rgb]{0.70,0.90,0.80} 60 & \cellcolor[rgb]{0.88,0.96,0.92} 75 & \cellcolor[rgb]{0.91,0.97,0.94} 78 & \cellcolor[rgb]{0.64,0.88,0.76} 54 & \cellcolor[rgb]{0.73,0.91,0.82} 62\\
Graphite &\cellcolor[rgb]{0.73,0.91,0.82} 62 & \cellcolor[rgb]{0.55,0.85,0.70} 46 & \cellcolor[rgb]{0.90,0.97,0.93} 77 & \cellcolor[rgb]{0.69,0.90,0.79} 58 & \cellcolor[rgb]{0.77,0.92,0.84} 65 & \cellcolor[rgb]{0.58,0.86,0.72} 48 & \cellcolor[rgb]{0.56,0.85,0.71} 47 & \cellcolor[rgb]{0.73,0.91,0.82} 62\\
Oil paint &\cellcolor[rgb]{0.58,0.86,0.72} 48 & \cellcolor[rgb]{0.46,0.82,0.64} 38 & \cellcolor[rgb]{0.51,0.84,0.67} 42 & \cellcolor[rgb]{0.96,0.99,0.97} 82 & \cellcolor[rgb]{0.44,0.81,0.63} 36 & \cellcolor[rgb]{0.64,0.88,0.76} 54 & \cellcolor[rgb]{0.71,0.90,0.80} 60 & \cellcolor[rgb]{0.80,0.93,0.87} 68\\
Pen ink &\cellcolor[rgb]{0.65,0.88,0.77} 55 & \cellcolor[rgb]{0.66,0.89,0.78} 56 & \cellcolor[rgb]{0.79,0.93,0.86} 67 & \cellcolor[rgb]{0.64,0.88,0.76} 54 & \cellcolor[rgb]{0.93,0.98,0.95} 79 & \cellcolor[rgb]{0.68,0.89,0.79} 58 & \cellcolor[rgb]{0.60,0.87,0.74} 51 & \cellcolor[rgb]{0.74,0.91,0.83} 63\\
Vector art &\cellcolor[rgb]{0.76,0.92,0.84} 65 & \cellcolor[rgb]{0.67,0.89,0.78} 56 & \cellcolor[rgb]{0.53,0.84,0.69} 44 & \cellcolor[rgb]{0.58,0.86,0.72} 48 & \cellcolor[rgb]{0.66,0.89,0.77} 55 & \cellcolor[rgb]{1.00,1.00,1.00} 86 & \cellcolor[rgb]{0.50,0.83,0.67} 41 & \cellcolor[rgb]{0.61,0.87,0.74} 51\\
Watercolor &\cellcolor[rgb]{0.61,0.87,0.74} 51 & \cellcolor[rgb]{0.56,0.85,0.71} 47 & \cellcolor[rgb]{0.71,0.90,0.81} 60 & \cellcolor[rgb]{0.88,0.96,0.92} 76 & \cellcolor[rgb]{0.69,0.90,0.79} 58 & \cellcolor[rgb]{0.66,0.89,0.78} 56 & \cellcolor[rgb]{0.80,0.93,0.87} 68 & \cellcolor[rgb]{0.77,0.92,0.85} 65\\
ImageNet &\cellcolor[rgb]{0.62,0.87,0.75} 52 & \cellcolor[rgb]{0.40,0.80,0.60} 32 & \cellcolor[rgb]{0.59,0.86,0.73} 50 & \cellcolor[rgb]{0.71,0.90,0.81} 60 & \cellcolor[rgb]{0.52,0.84,0.68} 43 & \cellcolor[rgb]{0.57,0.86,0.71} 48 & \cellcolor[rgb]{0.58,0.86,0.72} 48 & \cellcolor[rgb]{0.95,0.98,0.97} 82\\
\hline
$N-1$ Baseline &\cellcolor[rgb]{0.82,0.94,0.88} 70 & \cellcolor[rgb]{0.81,0.94,0.87} 69 & \cellcolor[rgb]{0.86,0.95,0.91} 73 & \cellcolor[rgb]{0.86,0.95,0.90} 73 & \cellcolor[rgb]{0.90,0.97,0.93} 77 & \cellcolor[rgb]{0.93,0.98,0.95} 79 & \cellcolor[rgb]{0.74,0.91,0.83} 63 & \cellcolor[rgb]{0.85,0.95,0.90} 73\\
\end{tabular}

}
\caption{\label{tab:gen-mtx}Domain transfer from one medium to another. This is mean object recognition performance when trained on a single artistic medium (row) and evaluated on a single medium (col). The ``$N-1$ Baseline'' model was trained on the other artistic media types.
}
\end{table}

%%%%% --> Moved to supplementary material
%
\subsection{Style and aesthetics}
Turning away from object categories for a moment, we now consider tasks related to stylistic information using the emotion and media labels in our dataset. We first investigate the effectiveness of different pre-trained features on emotion and media classification, and then show how to improve aesthetic and style classifiers on other artistic datasets.

\textbf{Feature comparison.}
How well can object recognition models transfer to emotion and media classification? Do models fine-tuned for style tasks forget their object recognition capabilities?
%Do they outperform StyleNet~\cite{Fang2015CollaborativeFL}, which is a CNN specialized to a style prediction task?
%{\color{red} This section is confusing. The main message is not clear to me, make the statement simpler and clearer. Also, the Fine-tuned StyleNet in Fig8 is not mentioned in text.}
%Here, we evaluate several different CNN features against our emotion and media labels.
To find out, we compare a linear SVM trained on pre-trained ResNet features to two style prediction models: a linear SVM trained on StyleNet features~\cite{Fang2015CollaborativeFL} and a StyleNet fine-tuned on Behance Artistic Media. The original StyleNet model was a GoogLeNet that was trained for a style prediction task. We hypothesize that it may outperform ResNet on tasks related to emotion and media classification.

\begin{figure}[t]
\includegraphics[width=0.49\linewidth]{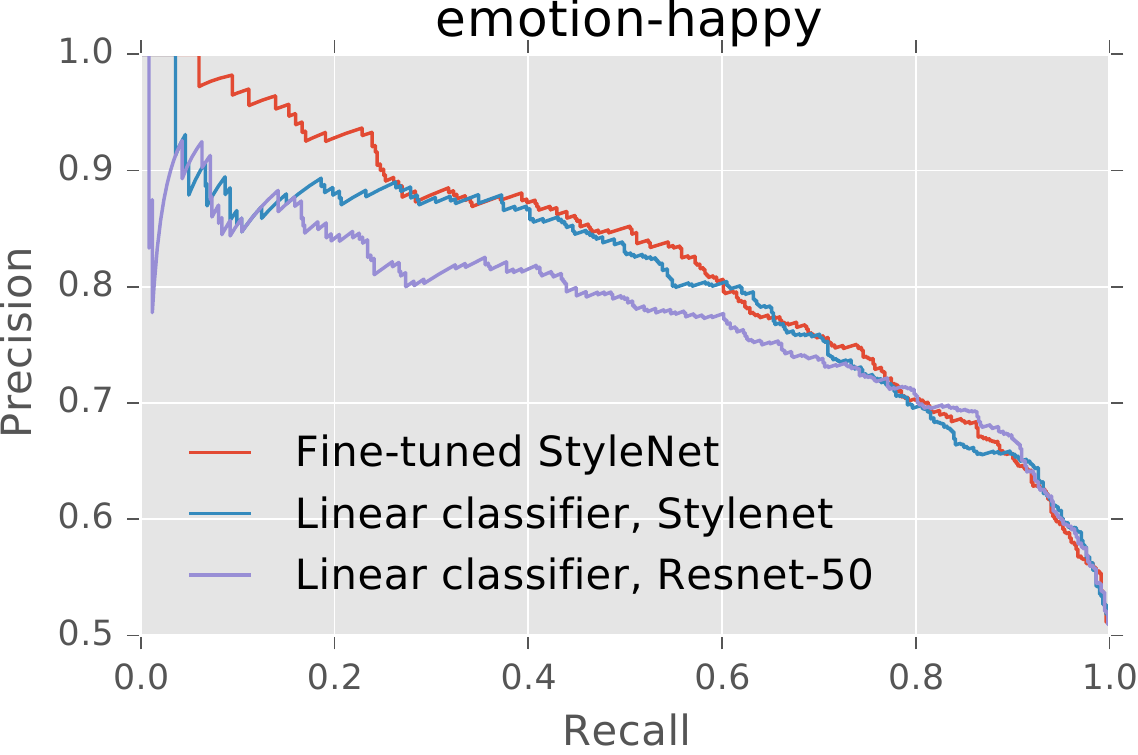}
\includegraphics[width=0.49\linewidth]{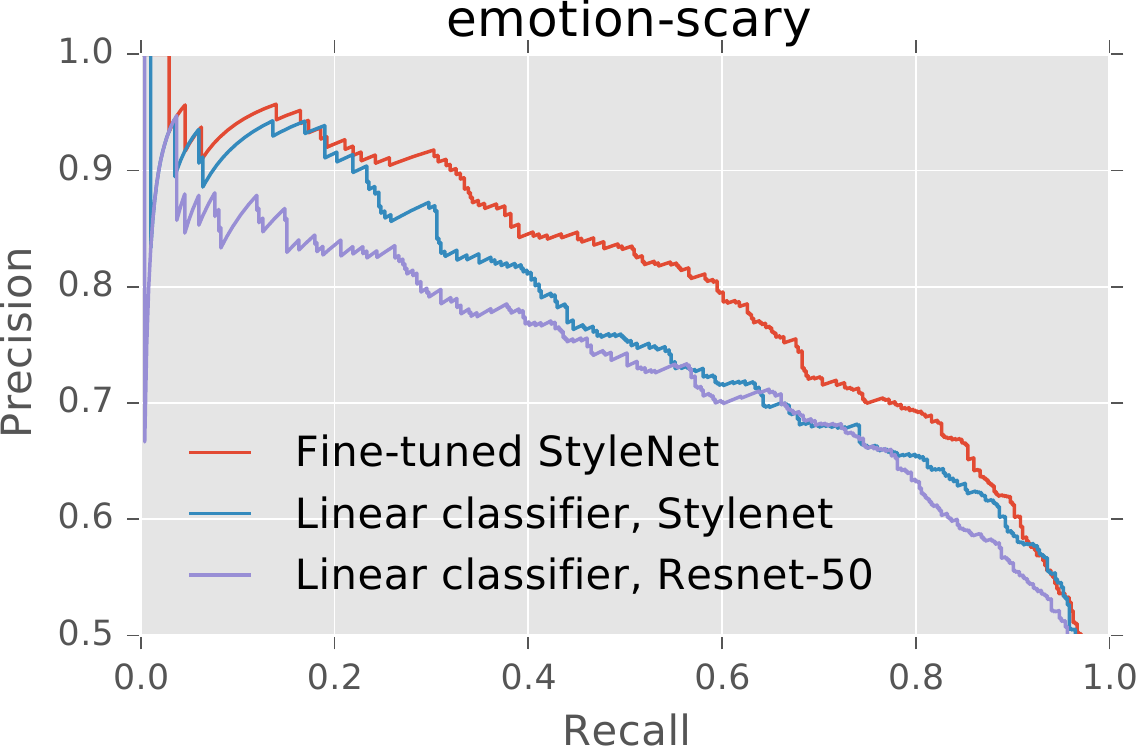}
\\
\includegraphics[width=0.49\linewidth]{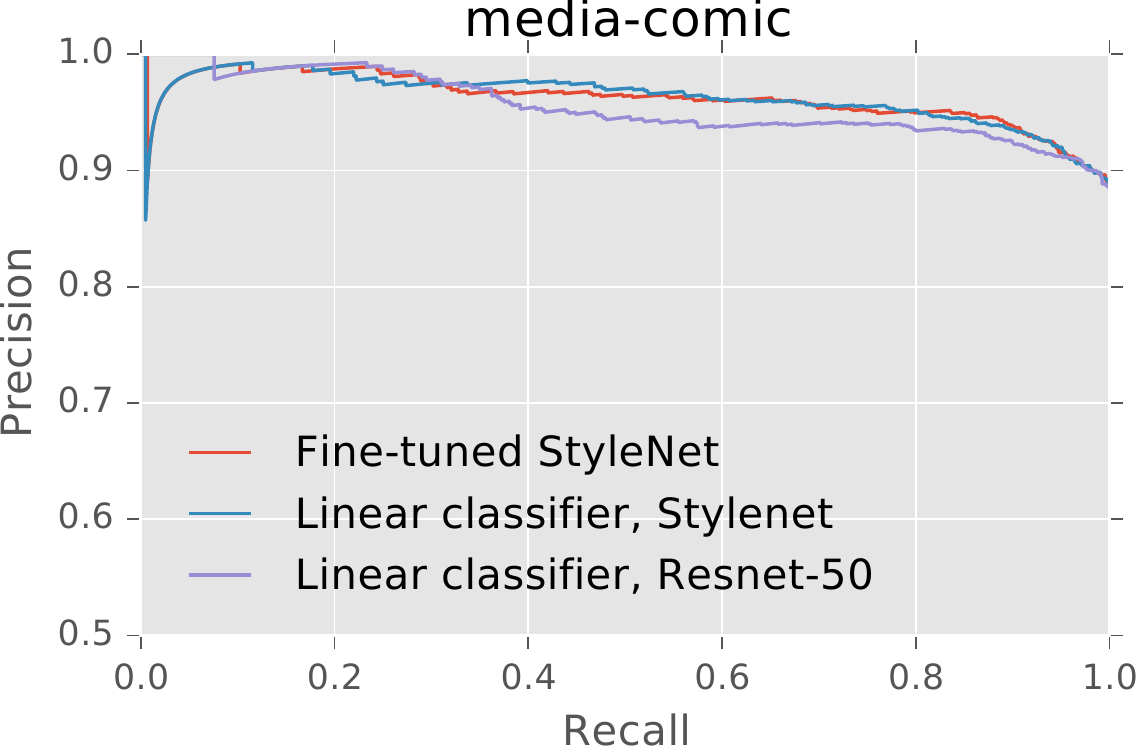}
\includegraphics[width=0.49\linewidth]{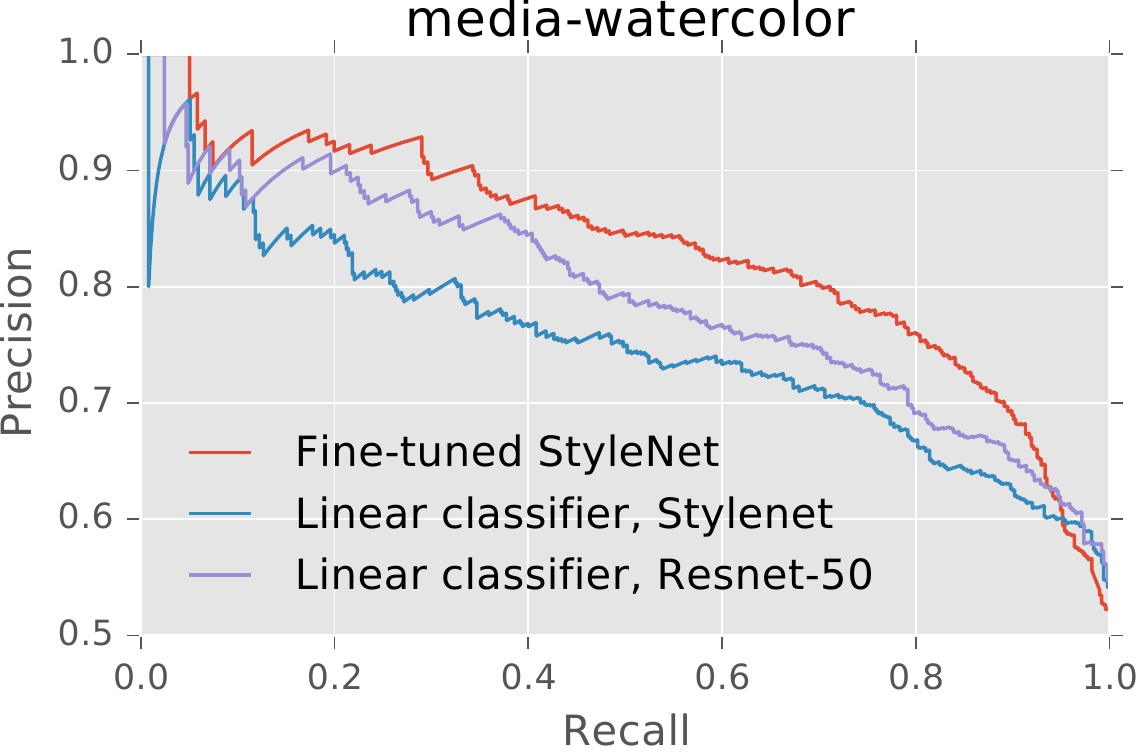}
\\
\includegraphics[width=0.49\linewidth]{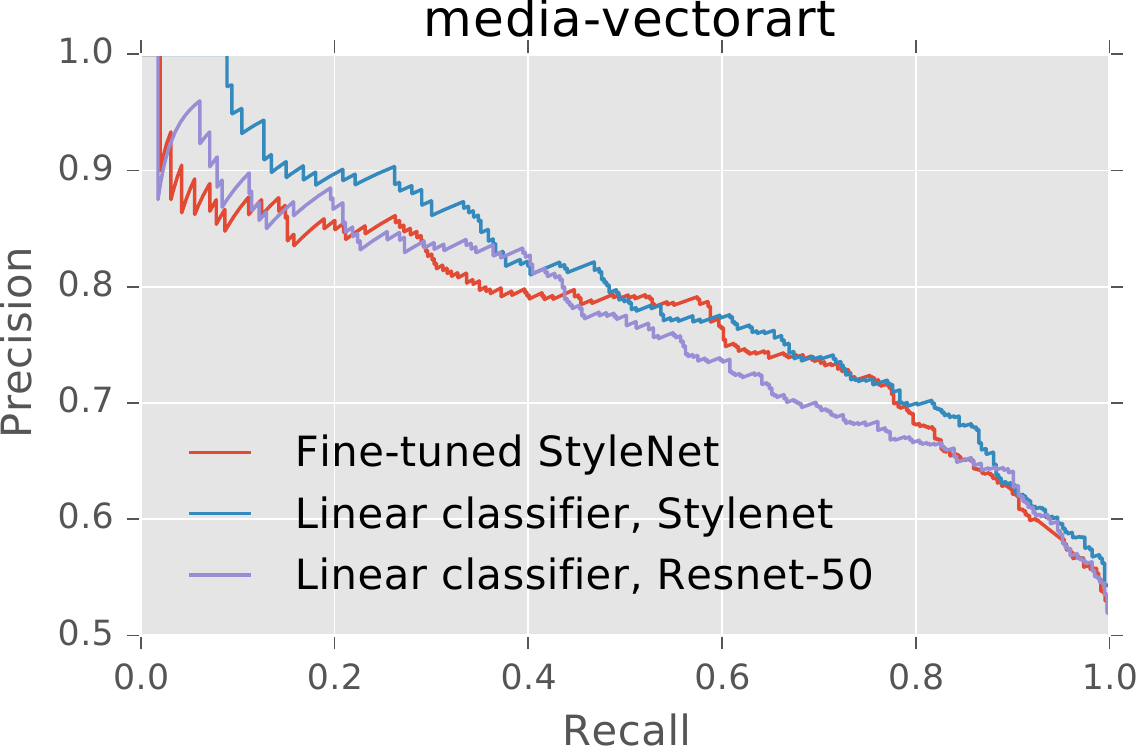}
\includegraphics[width=0.49\linewidth]{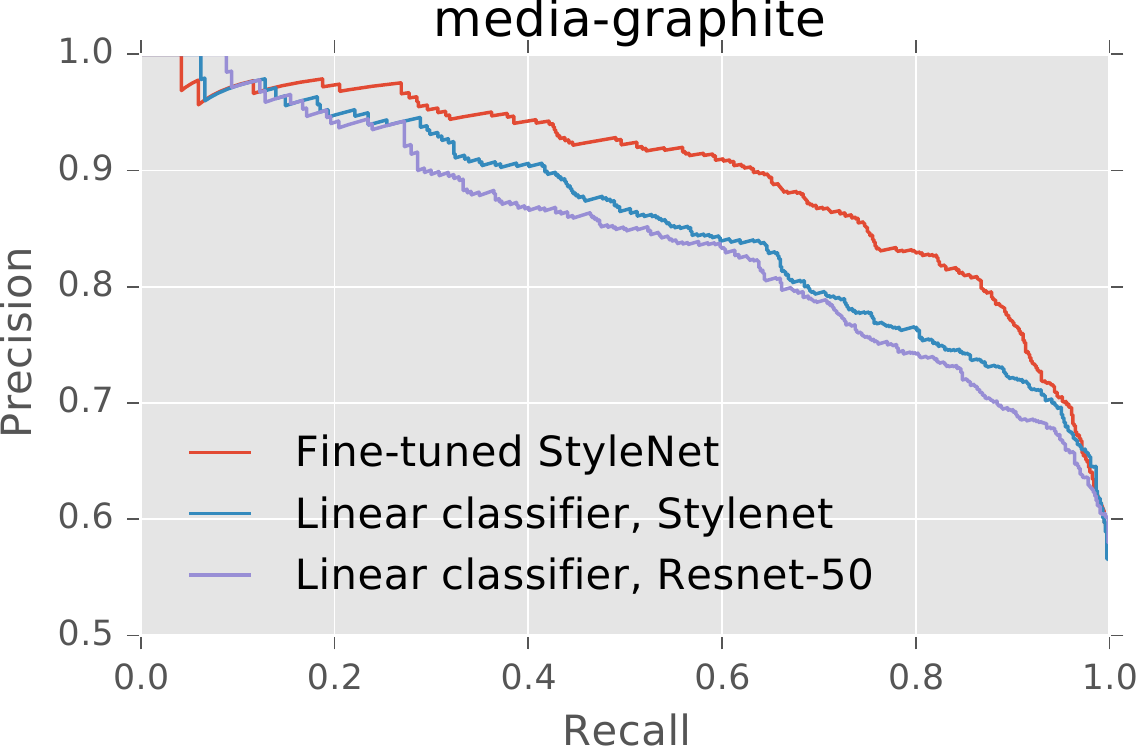}
\caption{\label{fig:style-feature-pr} Performance of different features on style attributes.}
\end{figure}

We evaluate these models on held-out human labels for each attribute.
%We think CNNs specialized to ImageNet should perform better on object attributes, while CNNs specialized on aesthetic tasks should perform better on style attributes. To test this hypothesis,
Performance for six attributes is shown in Fig.~\ref{fig:style-feature-pr}. For all four emotion attributes and 4/6 media attributes, the AP of linear classifiers on StyleNet features outperformed ImageNet-derived features. However, ImageNet-derived features have higher AP than StyleNet features on all nine content attributes.
%
%This supports the view that vision systems trained to distinguish object categories can more easily transfer domain knowledge to distinguishing artistic objects as well. One might wonder whether information about artistic style would be more important for this task, but that is not the case; instead,
%From this, we conclude that fine-tuning a network for style prediction tasks~\cite{Fang2015CollaborativeFL} makes it more suitable to distinguish emotion and media attributes at the cost of reducing object detection performance.
Different features are useful for content tasks compared to emotion/media tasks, and our dataset can help uncover these effects.

%
%
%\subsubsection{Using Behance Artistic Media to improve aesthetic classifiers on other datasets}
\label{sec:JAM}
\textbf{Aesthetic classification on other datasets.}
Other artistic datasets such as Wikipantings and AVA contain photographic style annotations. How well do models trained on our dataset perform on these datasets?
We show that automatic labels from Behance Artistic Media can slightly improve style classification on existing datasets. We evaluate on the three datasets introduced in~\cite{Karayev2014RecognizingIS}: 80,000 images in 20 photographic styles on Flickr, 85,000 images from the top 25 styles on Wikipaintings, and the 14,000 images with 14 photographic styles from the hand-labeled set of AVA~\cite{Murray2012AVAAL}. For comparison to previous work~\cite{Fang2015CollaborativeFL}, we report AVA classification accuracy calculated only on the 12,000 images that have a single style label. %Note that images from AVA may have more than one positive style label, so we report two metrics: the mean AP across each attribute as reported in~\cite{openimages} and a simpler accuracy score similar to~\cite{Fang2015CollaborativeFL} calculated only on the 12,000 images that have a single style label.

To solve this task, we train a joint attribute model (\emph{JAM}) that outputs all attribute scores simultaneously. Each training sample $(x, i, \ell)$ is a tuple of image $x$, attribute index $i$, and label $\ell \in \{-1, 1\}$. It is not suitable to train this model using ordinary cross entropy because each attribute is not mutually exclusive. Thus, we must use a loss function with two properties: each attribute output should be independent of other attributes and unknown attribute values should not induce any gradient. We lift image $x$ to a 20-dimensional partial attribute vector $\hat{y} \in \mathcal{R}^{20}$, where $\hat{y}_{j \neq i} = 0$ and $\hat{y}_{j = i} = \ell$. This allows us to train using a soft-margin criterion,
$loss(x, y) = \frac{1}{20} \sum_i \log(1 + \exp(- \hat{y}_i y_i)).$
Our JAM model is a fine-tuned ResNet-50 model with a linear projection from 1,000 to 20 dimensions. % Except for the last layer, they share the same architecture and are merely trained using different loss functions.
We trained our model for 100 epochs, starting with a learning rate of 0.1 and multiplying it by 0.93 every epoch. The training set includes roughly 2 million images evenly sampled between attributes and evenly distributed between positive and negative images drawn from the automatically-labeled images in Behance Artistic Media.

Results are shown on Table~\ref{tab:joint-model}. On all three challenges, our model shows improved results compared to both the original ResNet-50 and StyleNet. This shows that Behance imagery is rich and diverse enough to improve style recognition tasks on other datasets. This is particularly interesting because Flickr and AVA are both focused on photographic style. Categories in AVA are chosen to be useful for aesthetic quality prediction tasks.
%In a sense, we have shown that a model's knowledge of emotions and media could potentially transfer to photographic style and aesthetic prediction.
This shows that models can train on our dataset to improve performance on other aesthetic classification datasets.

\begin{table}[t]
  \centering
  \scalebox{0.8}{
  \begin{tabular}{l|rrr}
\hline
& JAM & ResNet-50\vspace{-0.4em} & StyleNet~\cite{Fang2015CollaborativeFL} \\
&&{\scriptsize (ImageNet)}&\\
\hline
Flickr & \textbf{0.389} & 0.376 & 0.372 \\
Wikipaintings & \textbf{0.508} & 0.505 & 0.414 \\
AVA & \textbf{0.615} & 0.603 & 0.560\\
\hline
  \end{tabular}
  }
  \caption{\label{tab:joint-model}Performance of our joint model for style detection on other datasets}
\end{table}

%\subsection{Visual subspace learning for task-specific image retrieval}
\subsection{Visual subspace learning}
Finally, we conclude by showing how to learn task-specific subspaces to retrieve images according to content, emotion, or media similarity.
One disadvantage of the joint attribute model mentioned above is the lack of separate feature spaces for each task. Consider the case of image retrieval where the goal is to find images that share the content of a query image but not necessarily its artistic medium. We can use Behance Artistic Media to solve this task by treating it as a visual subspace learning problem. Starting from a pre-trained ResNet shared representation, we remove the top layer and add three branches for content, emotion, and media. Each branch contains a linear projection down to a 64-dimensional subspace and a final projection down to label space. The final model is trained similarly to the model in Sec.~\ref{sec:JAM}. Only the initial ResNet weights are shared; the embedding is separate for each task. We qualitatively show three images close to the query within each task-specific embedding.

The results show that this simple strategy can learn sensible task-specific embeddings. Neighbors in latent-content space generally match the content of the query and neighbors in latent-media space generally match the query's artistic medium. The effect is qualitatively weaker for emotion space, perhaps because of the limited label set. From a human inspection of 100 random queries, the precision-at-10 for content, media, and emotion is 0.71, 0.91, and 0.84 respectively. Media and emotion precision-at-10 are slightly improved compared to our shared feature baseline of 0.80, 0.87, 0.80, which could be explained if the shared representation focuses almost exclusively on content.
One limitation of this approach is that without any conditioning, the three learned subspaces tend to be correlated: objects close in media-space or emotion-space sometimes share content similarity. Our dataset could provide a rich resource for feature disentangling research.

%\todo Also, you haven't mentioned the potential of doing feature disentangling. I think we can weakly mention it somewhere.

\begin{figure}[t]
  \centering
  \includegraphics[width=0.9\linewidth]{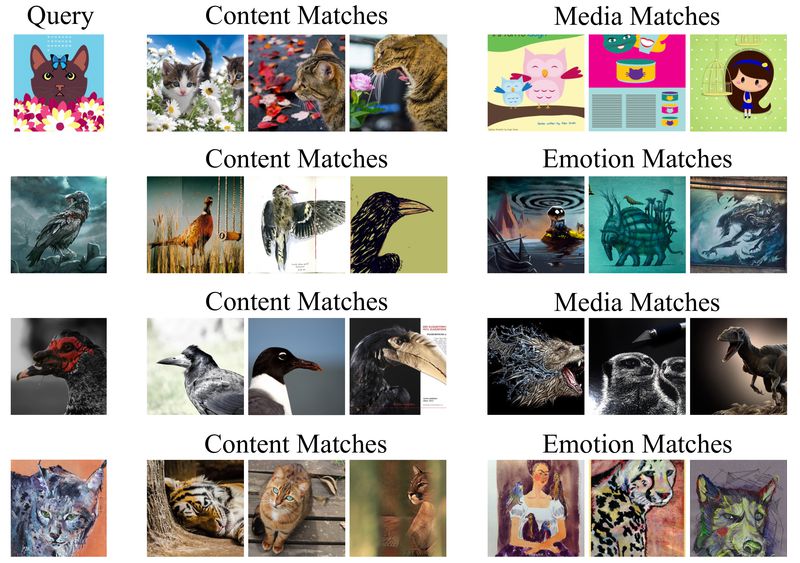}
  \caption{\label{fig:disentangling-neighbors}Retrieval results showing a query image, three content neighbors, and three neighbors from another facet}
\end{figure}

\subsection{Visualizing the learned model}

\begin{figure}[t!]
\centering
\includegraphics[width=1.0\linewidth,height=4.5cm]{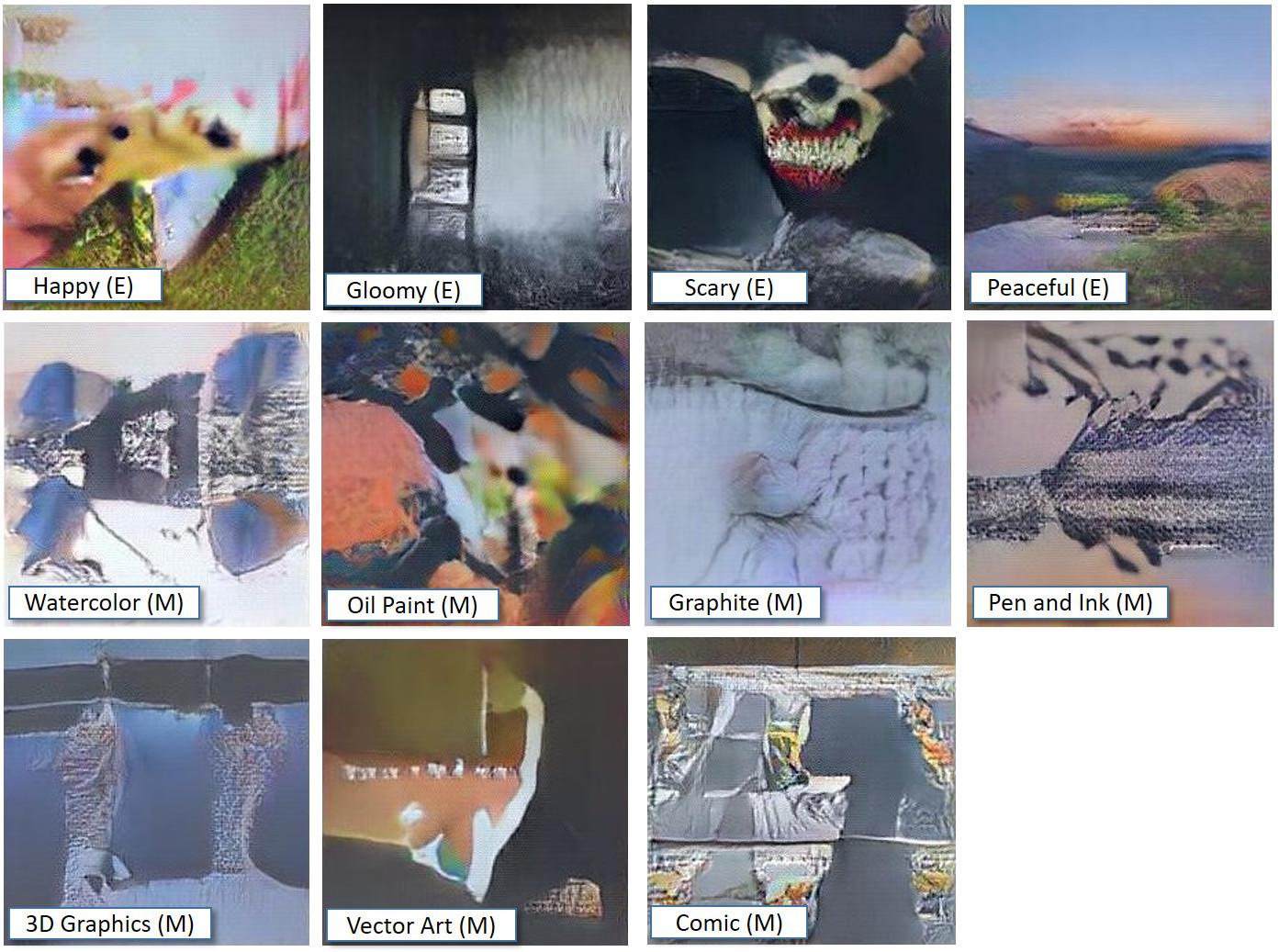}
   \caption{Visualizing learning: maximum activations for each of the media and emotion styles.\label{fig:vizstyles}
}
\end{figure}

We qualitatively explore the kind of visual style cues learnable from the proposed dataset in Fig.~\ref{fig:vizstyles}.  A dataset of ~110k images was formed by sorting all 65m Behance Artistic Media images by likelihood score for each of the 7 media and 4 emotion attributes, and sampling the top 10k images in each case.  Duplicate images selected across attributes were discarded.   A modified Alexnet\cite{Krizhevsky2012} (fc6 layer 1024-D, fc7 layer 256-D) was trained from scratch on the 11 style (media and emotion) attributes for 40 epochs via SGD with learning rate 0.01.  Nguyen \etal \cite{Nguyen2016} recently proposed a deep generator network (DGN) based visualization technique for synthesizing stimuli preferred by neurons through combination of a truncated (ImageNet trained) CaffeNet and up-convolutional network initialized via white noise.  We run the DGM-AM variant of their process for 200 iterations, using a learning rate of 2.0 and weighting factor 99. The images synthesized for several media types (\eg {\em graphite}, {\em oil-paint} and {\em watercolor} paintings) epitomize textures commonly encountered in these art forms although styles exhibiting structural combination of flatter regions are less recognizable.  Fragments of objects commonly recognizable within emotion-based styles (e.g. teeth for {\em scary}, bleak windows in {\em gloomy} or landscapes in {\em peaceful} are readily apparent.  %Inspecting the 10k training images for each classes, there is clear semantic prior in the emotion-based styles (for examples, almost 80\% of peaceful images contain trees and water) reflected in the visualizations.

\section{Conclusion}
%\todo {\color{red} Update the conclusion to incorporate new experiments, i.e., domain transfer, subspace learning (or feature disentangling), visual retrieval)}
Computer vision systems need not be constrained to the domain of photography.
%Here, we show how the rich field of artistic imagery can benefit machine vision systems.
%
We propose a new dataset, ``Behance Artistic Media'' (\emph{BAM!}), a repository of millions of images posted by professional and commercial artists representing a broad snapshot of contemporary artwork. We collected a rich vocabulary of emotion, media, and content attributes that are visually distinctive and representative of the diversity found in Behance.

However, though Behance does include tag metadata, we showed that these tags are too noisy to learn directly. Further, the scale of Behance makes brute-force crowdsourcing unattractive.
To surmount these issues, we collected labels via a hybrid human-in-the-loop system that uses deep learning to amplify human annotation effort while meeting desired quality guarantees. % This allows existing machine vision systems to focus crowd attention on the images that need human expertise. Our annotation pipeline collects labels at a fraction of the cost of brute force labeling while meeting precision/recall guarantees of our choosing.

The resulting dataset is useful for several computer vision tasks.
We use it to highlight the representation gap of current object detection systems trained on photography, showing that Behance captures a wider gamut of representation styles than current sets such as VOC and ImageNet. Different artistic media in Behance have unique aesthetics, providing an interesting test bed for domain transfer tasks, and different features prove useful for content tasks compared to media/emotion classification.
We also use Behance to improve the performance of style classification on other datasets, showing that researchers can train on our dataset for a marked improvement in performance. Finally, we conclude with a subspace learning task for retrieving images based on their content or artistic media.

% - The resulting dataset is useful for several computer vision tasks.
%   - We can use it to highlight the representation gap of current object detection systems
%   - We can use it to improve the performance of existing style classifiers on other datasets

We believe our dataset provides a good foundation for further research into the underexplored realm of large-scale artistic imagery.

\section*{Acknowledgments}
This work is partly funded by an NSF Graduate Research Fellowship award (NSF DGE-1144153, Author 1), a Google Focused Research award (Author 6), a Facebook equipment donation to Cornell University, and Adobe Research.

All images in this paper were shared with Creative Commons attribution licenses and we thank the creators for sharing them:
-jose-, AARON777, AbrahamCruz, AndreaRenosto, AndrewLili, AngelDecuir, AnnaOfsa, Arianne, AstridCarolina, Aubele, ChumaMartin, DannyWu, DubrocDesign, Elfimova, EtienneBas, ExoticMedia, Folkensio, GinoBruni, GloriaFl, HattyEberlein, Hectorag, ImaginaryFS, IsmaelCG, J\_Alfredo, JaclynSovern, JanayFrazier, JaneJake, JodiChoo, Joshua\_Ernesto, KimmoGrabherr, Kiyorox, Marielazuniga, Mirania, NIELL, NikolayDimchev, NourNouralla, OlivierPi, Ooldigital, Pleachflies, PranatiKhannaDesign, PraveenTrivedi, Radiotec1000, RamshaShakeel, RaoulVega, Roxxo2, Royznx, RozMogani, SaanaHellsten, Sanskruta, Slavina\_InkA, SwannSmith, TCaetano, THE-NEW, Thiagoakira, TomGrillo, WayneMiller, Yipori, ZaraBuyukliiska, adovillustration, alobuloe61b, amos, animatorcreator, answijnberg, anthonyrquigley, antoniovizzano, apedosmil, artofmyn, asariotaki, asjabor, assaadawad, atanaschopski, averysauer, ballan, bartosz, beissola, bilbo3d, bluelaky, brainlessrider, brandplus, brittany-naundorff, bucz, campovisual, candlerenglish, carlosmessias, carlosoporras, castilloseas, cgshotgun, charbelvanille, charlottestrawbridge, chin2off, cknara, coyote-spark, crazydiamondstudio, cristianbw, crivellaro, cube74, customshoot, daconde, dafrankish, danielstarrason, dariomaggiore, deboralstewart, dedos, designrahul, digital-infusion, ememonkeybrain, esanaukkarinen, evanwitek, fabsta, flochau, florencia\_agra, foundationrugs, gatomontes, giuseppecristiano, gumismydrug, gurudesigner, heroesdesign, higoszi, ifyouprefer, ikuchev, illustrationdesign, irf, irwingb, ivancalovic, jabaarte, jakobjames, janaramos, janeschmittibarra, jdana, jota6six, juliencoquentin, julienps, justindrusso, kalalinks, karoliensoete, katesimpressions, katiemott, kpkeane, larawillson, leovelasco, leturk, lindseyrachael, lisa\_smirnova, littleriten, manf\_v, margotsztajman, martabellvehi, matildedigmann, maviemamuse, mechiparra, michaelkennedy, monroyilustrador, mvitacca, nataliamon, nati24k, nazpa, nickelcurry, nicopregelj, nobleone, nspanos, nunopereira, olaolaolao, olivierventura, ollynicholson, omarmongepinal, onthebrinkartstudios, osamaaboelezz, pamelapg, paperbenchdesign, parfenov, pengyangjun, pilarcorreia, pilot4ik, polyesterdress, priches, pumaroo, ragamuffin, rainingsuppie, raquelirene, rasgaltwc, rebeccahan, redric, rikishi, roosariski, rrabbstyjke, rufusmediateam, sairarehman, saketattoocrew1, samii69, sandhop, santarromana, santocarone, sashaanikusko, scottchandler, sebagresti, sehriyar, shaneferns, shaungmakeupartist, siqness, snowballstudio, softpill, sompson, sophieecheetham, starkdesigns, studiotuesday, tadzik, tatianaAdz, taylorboren, taylored, tensil, theCreativeBarn, tomtom\_ungor, toniitkonen, travisany, tweek, ukkrid, visual3sesenta, zarifeozcalik, zmahmoud, zwetkow

{\small
\bibliographystyle{ieee}
\bibliography{egbib}

\begin{thebibliography}{10}\itemsep=-1pt

\bibitem{Borth2013}
D.~Borth, R.~Ji, T.~Chen, T.~Breuel, and S.-F. Chang.
\newblock {Large-scale visual sentiment ontology and detectors using adjective
  noun pairs}.
\newblock In {\em Proc.~MM}, 2013.

\bibitem{Crowley14a}
E.~J. Crowley and A.~Zisserman.
\newblock In search of art.
\newblock In {\em Workshop on Computer Vision for Art Analysis, ECCV}, 2014.

\bibitem{Csurka2017DomainAF}
G.~Csurka.
\newblock Domain adaptation for visual applications: A comprehensive survey.
\newblock {\em CoRR}, abs/1702.05374, 2017.

\bibitem{Cui2015FinegrainedCA}
Y.~Cui, F.~Zhou, Y.~Lin, and S.~J. Belongie.
\newblock Fine-grained categorization and dataset bootstrapping using deep
  metric learning with humans in the loop.
\newblock In {\em Proc.~CVPR}, 2016.

\bibitem{Dhar2011HighLD}
S.~Dhar, V.~Ordonez, and T.~L. Berg.
\newblock High level describable attributes for predicting aesthetics and
  interestingness.
\newblock In {\em CVPR}, 2011.

\bibitem{Fang2015CollaborativeFL}
C.~Fang, H.~Jin, J.~Yang, and Z.~L. Lin.
\newblock Collaborative feature learning from social media.
\newblock {\em CoRR}, abs/1502.01423, 2015.

\bibitem{Farhadi2009}
A.~Farhadi, I.~Endres, D.~Hoiem, and D.~Forsyth.
\newblock {Describing objects by their attributes}.
\newblock In {\em CVPR Workshops 2009}, pages 1778--1785, 2009.

\bibitem{Gatys2015}
L.~A. Gatys, A.~S. Ecker, and M.~Bethge.
\newblock Image {S}tyle {T}ransfer {U}sing {C}onvolutional {N}eural {N}etworks.
\newblock In {\em Proc.~CVPR}, 2016.

\bibitem{Ginosar2015}
S.~Ginosar, D.~Haas, T.~Brown, and J.~Malik.
\newblock {Detecting people in cubist art}.
\newblock In {\em Lecture Notes in Computer Science (including subseries
  Lecture Notes in Artificial Intelligence and Lecture Notes in
  Bioinformatics)}, volume 8925, pages 101--116, 2015.

\bibitem{DBLP:journals/corr/HeZRS15}
K.~He, X.~Zhang, S.~Ren, and J.~Sun.
\newblock Deep residual learning for image recognition.
\newblock In {\em Proc.~CVPR}, 2016.

\bibitem{Izadinia2015}
H.~Izadinia, B.~C. Russell, A.~Farhadi, M.~D. Hoffman, and A.~Hertzmann.
\newblock {Deep Classifiers from Image Tags in the Wild}.
\newblock In {\em Proc.~Multimedia COMMONS}, 2015.

\bibitem{Jou2015}
B.~Jou, T.~Chen, N.~Pappas, M.~Redi, M.~Topkara, and S.-F. Chang.
\newblock {Visual Affect Around the World: A Large-scale Multilingual Visual
  Sentiment Ontology}.
\newblock In {\em Proc.~MM}, pages 159--168, 2015.

\bibitem{Karayev2014RecognizingIS}
S.~Karayev, M.~Trentacoste, H.~Han, A.~Agarwala, T.~Darrell, A.~Hertzmann, and
  H.~Winnem\"{o}ller.
\newblock Recognizing image style.
\newblock In {\em Proc.~BMVC}, 2014.

\bibitem{openimages}
I.~Krasin, T.~Duerig, N.~Alldrin, A.~Veit, S.~Abu-El-Haija, S.~Belongie,
  D.~Cai, Z.~Feng, V.~Ferrari, V.~Gomes, A.~Gupta, D.~Narayanan, C.~Sun,
  G.~Chechik, and K.~Murphy.
\newblock Openimages: A public dataset for large-scale multi-label and
  multi-class image classification.
\newblock {\em Dataset available from https://github.com/openimages}, 2016.

\bibitem{Krause2016TheUE}
J.~Krause, B.~Sapp, A.~Howard, H.~Zhou, A.~Toshev, T.~Duerig, J.~Philbin, and
  L.~Fei-Fei.
\newblock The unreasonable effectiveness of noisy data for fine-grained
  recognition.
\newblock In {\em Proc.~ECCV}, 2016.

\bibitem{Krizhevsky2012}
A.~Krizhevsky, I.~Sutskever, and G.~Hinton.
\newblock Imagenet classification with deep convolutional neural networks.
\newblock In {\em Proc. NIPS}, 2012.

\bibitem{mscoco}
T.~Y. Lin, M.~Maire, S.~Belongie, J.~Hays, P.~Perona, D.~Ramanan,
  P.~Doll{\'{a}}r, and C.~L. Zitnick.
\newblock {Microsoft COCO: Common objects in context}.
\newblock In {\em Proc.~ECCV}, pages 740--755, 2014.

\bibitem{Lin_2016_CVPR}
T.-Y. Lin and S.~Maji.
\newblock Visualizing and understanding deep texture representations.
\newblock In {\em The IEEE Conference on Computer Vision and Pattern
  Recognition (CVPR)}, June 2016.

\bibitem{Liu2015}
W.~Liu, D.~Anguelov, D.~Erhan, C.~Szegedy, S.~Reed, and A.~C. Berg.
\newblock {SSD: Single Shot MultiBox Detector}.
\newblock In {\em Proc.~ECCV}, 2016.

\bibitem{MisraNoisy16}
I.~Misra, C.~L. Zitnick, M.~Mitchell, and R.~Girshick.
\newblock {Seeing through the Human Reporting Bias: Visual Classifiers from
  Noisy Human-Centric Labels}.
\newblock In {\em CVPR}, 2016.

\bibitem{Murray2012AVAAL}
N.~Murray, L.~Marchesotti, and F.~Perronnin.
\newblock {AVA}: {A} large-scale database for aesthetic visual analysis.
\newblock In {\em Proc.~CVPR}, 2012.

\bibitem{Nguyen2016}
A.~Nguyen, A.~Dosovitskiy, J.~Yosinski, T.~Brow, and J.~Clune.
\newblock Synthesising the preferred inputs for neurons in neural networks via
  deep generator networks.
\newblock In {\em Proc. NIPS}. IEEE, 2016.

\bibitem{Obrador2012TowardsCA}
P.~Obrador, M.~A. Saad, P.~Suryanarayan, and N.~Oliver.
\newblock Towards category-based aesthetic models of photographs.
\newblock In {\em Proc.~MMM}, 2012.

\bibitem{7078994}
V.~M. Patel, R.~Gopalan, R.~Li, and R.~Chellappa.
\newblock Visual domain adaptation: A survey of recent advances.
\newblock {\em IEEE Signal Processing Magazine}, 32(3):53--69, May 2015.

\bibitem{Patterson2012SUNAD}
G.~Patterson and J.~Hays.
\newblock Sun attribute database: Discovering, annotating, and recognizing
  scene attributes.
\newblock In {\em Proc.~CVPR}, 2012.

\bibitem{peng2016}
K.-C. Peng, A.~Sadovnik, A.~Gallagher, and T.~Chen.
\newblock Where do emotions come from? predicting the emotion stimuli map.
\newblock In {\em Proc.~ICIP}, 2016.

\bibitem{Plutchik2001}
R.~Plutchik.
\newblock {The nature of emotions: Human emotions have deep evolutionary
  roots}.
\newblock {\em American Scientist}, 89(4):344--350, 2001.

\bibitem{NunoSemanticMultinomials}
N.~Rasiwasia, P.~J. Moreno, and N.~Vasconcelos.
\newblock Bridging the gap: Query by semantic example.
\newblock {\em IEEE Trans. Multimedia}, 9:923--938, 2007.

\bibitem{Redmon2016}
J.~Redmon, S.~Divvala, R.~Girshick, and A.~Farhadi.
\newblock {You Only Look Once: Unified, Real-Time Object Detection}.
\newblock In {\em CVPR 2016}, pages 779--788, 2016.

\bibitem{DBLP:journals/corr/SzegedyLJSRAEVR14}
C.~Szegedy, W.~Liu, Y.~Jia, P.~Sermanet, S.~E. Reed, D.~Anguelov, D.~Erhan,
  V.~Vanhoucke, and A.~Rabinovich.
\newblock Going deeper with convolutions.
\newblock In {\em Proc.~CVPR}, 2016.

\bibitem{You2016BuildingAL}
Q.~You, J.~Luo, H.~Jin, and J.~Yang.
\newblock Building a large scale dataset for image emotion recognition: The
  fine print and the benchmark.
\newblock {\em CoRR}, abs/1605.02677, 2016.

\bibitem{Yu2015LSUNCO}
F.~Yu, Y.~Zhang, S.~Song, A.~Seff, and J.~Xiao.
\newblock Lsun: Construction of a large-scale image dataset using deep learning
  with humans in the loop, 2015.
\newblock arXiv:1506.03365.

\end{thebibliography}
}

% \section{\todo More experiments we could run}
% \begin{itemize}
% \item Correlation between content and style (E6)
% \item Search example images
% \end{itemize}

% See https://docs.google.com/presentation/d/1HkPmKNkiaPR06vjK0SX11QCxxW45JE6i7lO1NE7yg3E/edit#slide=id.g13916de101_0_82

\end{document}